\documentclass[10pt,journal,compsoc]{IEEEtran}

\ifCLASSOPTIONcompsoc
  \usepackage[nocompress]{cite}
\else
  \usepackage{cite}
\fi
\usepackage{url}
\usepackage{amsmath,amssymb,amsfonts}
\usepackage{extarrows}
\usepackage[pdftex]{graphicx}
\usepackage{hyperref}       % hyperlinks
\usepackage{xcolor}
\usepackage{subfigure}
\ifCLASSOPTIONcompsoc
\else
\fi

\usepackage[noend]{algpseudocode}
\usepackage{algorithm}
\usepackage{algorithmicx}
\usepackage{xspace}
\usepackage{dsfont}
\usepackage{epsfig}
\usepackage{epstopdf}
\usepackage{indentfirst}
\usepackage{bm}
\usepackage{color}
\usepackage{colortbl}
\usepackage{multirow}
\usepackage{ntheorem}
\usepackage{booktabs} % for nice tables
\usepackage{tablefootnote}
\usepackage{threeparttable} % for table notes
\usepackage{enumitem} % for nice itemize

\theoremheaderfont{\normalfont\bfseries}
\theorembodyfont{\itshape}
\theoremseparator{.}
\newtheorem{theorem}{Theorem}
\newtheorem{lemma}{Lemma}
\renewcommand{\IEEEQED}{\IEEEQEDopen}
\renewcommand{\IEEEproofindentspace}{0pt}

\newtheorem{mydef}{Definition}

\newcommand{\paratitle}[1]{\vspace{1.2ex}\noindent\textbf{#1}}
\newcommand{\ie}{\emph{i.e., }}
\newcommand{\aka}{\emph{a.k.a., }}
\newcommand{\eg}{\emph{e.g., }}

\newcommand{\etc}{\emph{etc}}
\newcommand{\wrt}{\emph{w.r.t. }}

\newcommand{\model}{STGDL\xspace}
\newcommand{\equ}{Eq.\xspace}
\newcommand{\fig}{Fig.\xspace}
\newcommand{\tab}{Tab.\xspace}

\newcommand{\thm}{Thm.\xspace}
\newcommand{\apx}{Appx.\xspace}

\newcommand{\code}{\url{https://github.com/bigscity/STGDL}}

\DeclareMathOperator{\st}{s.t.}

\begin{document}
% \title{Graph Decomposition to Facilitate Traffic Prediction}
% \title{Spatio-Temporal Graph Decomposition Learning for Multi-Factor Traffic Prediction}
% \title{Multi-Factor Spatio-Temporal Prediction\\ based on Graph Decomposition Learning:\\Theories and Models}
\title{Multi-Factor Spatio-Temporal Prediction\\ based on Graph Decomposition Learning}

\author{Jiahao Ji,  
        Jingyuan Wang,
        Yu Mou,
        and Cheng Long 
        % and Zhenhe Wu
\IEEEcompsocitemizethanks{
\IEEEcompsocthanksitem J. Ji, Y. Mou, and Z. Wu are with the School of Computer Science and Engineering, Beihang University, Beijing, China.
% \protect\\
% E-mail: \{jiahaoji, xx, xx\}@buaa.edu.cn. 
\IEEEcompsocthanksitem J. Wang is with the School of Computer Science and Engineering and School of Economics and Management, Beihang University, Beijing, China.
% \protect\\
% E-mail: jywang@buaa.edu.cn.
\IEEEcompsocthanksitem C. Long is with the School of Computer Science and Engineering, Nanyang Technological University, Singapore.
\IEEEcompsocthanksitem Corresponding author: jywang@buaa.edu.cn
}% <-this % stops an unwanted space
%\thanks{xxx}
}

\markboth{IEEE TRANSACTIONS ON KNOWLEDGE AND DATA ENGINEERING}%
{IEEE TRANSACTIONS ON KNOWLEDGE AND DATA ENGINEERING}

\IEEEtitleabstractindextext{%
\begin{abstract}
Spatio-temporal (ST) prediction is an important and widely used technique in data mining and analytics, especially for ST data in urban systems such as transportation data. In practice, the ST data generation process is usually affected by a mixture of multiple latent factors related to natural phenomena or human socioeconomic activities. Different factors generally affect some spatial areas but not all of them. However, existing ST prediction methods usually do not refine the impacts of different factors, but directly model the entangled impacts of multiple factors. This can increase the modeling difficulty of ST data and hurt the model interpretability at the same time. To this end, we propose a multi-factor spatio-temporal prediction task that predicts the evolution of the partial ST data under different factors separately and then combines them to produce a final overall prediction. We make two contributions to this task: an effective theoretical solution and a portable instantiation framework. Specifically, we first propose a theoretical solution called decomposed prediction strategy and prove its effectiveness from the perspective of information entropy theory. On top of that, we instantiate a novel model-agnostic framework, named spatio-temporal graph decomposition learning (STGDL), for multi-factor ST prediction. The framework consists of two main components: an automatic graph decomposition module that decomposes the original graph structure inherent in ST data into several subgraphs corresponding to different factors, and a decomposed learning network that learns the partial ST data on each subgraph separately and integrates them for the final prediction. The framework is theoretically guaranteed to reduce the prediction error and enhance the interpretability of predicted ST data. We conduct extensive experiments on four real-world ST datasets of two types of graphs, i.e., grid graph and network graph. Results show that our portable framework significantly reduces prediction errors of various ST models by 9.41\% on average (35.36\% at most). Furthermore, a case study reveals the interpretability potential of our framework as well as its predictions. Our code is at \code.
\end{abstract}

% In ST data generation, multiple latent factors linked to natural phenomena or human activities often interplay. These factors tend to affect specific spatial regions selectively. However, prevalent approaches typically do not distinguish between these factors, instead directly modeling their entangled effects. This can heighten the complexity of modeling ST data and compromise interpretability. To this end, we propose a multi-factor ST prediction task that forecasts partial data under distinct factors before consolidating them for a holistic prediction. We make two contributions to this task. Firstly, we propose a theoretical solution called decomposed prediction strategy and prove its effectiveness from the perspective of information entropy theory. Secondly, on top of that, we instantiate a novel portable framework named ST graph decomposition learning (STGDL). It consists of two components: an automatic graph decomposition module that decomposes the original graph structure into several subgraphs corresponding to different factors, and a decomposed learning network that learns partial ST data on every subgraph separately and integrates them for final predictions. Extensive experiments on four datasets show that portable STGDL reduces prediction errors of various ST models by 9.41\% on average (35.36\% at most). A case study further reveals the ability of STGDL to enhance prediction interpretability.

}

% Latent Multi-Factor Modeling

\maketitle

\IEEEpeerreviewmaketitle

\IEEEraisesectionheading{\section{Introduction}\label{sec:introduction}}

\IEEEPARstart{W}{ith} the ubiquitous use of GPS-enabled mobile devices and sensors, a huge volume of spatio-temporal (ST) data is emerging from a variety of domains, \eg urban transportation, meteorology, and public health. Through ST prediction techniques~\cite{wang2023explore, xie2023spatio, pan2022spatio, guo2021learning}, the ST data can support a growing number of applications ranging from individual smart mobility~\cite{zhang2022beyond} to public safety management~\cite{ji2022precision}. For example, ST traffic prediction, which aims to accurately forecast future traffic conditions from past traffic observations, can promote environmentally friendly commuting through bike-sharing initiatives~\cite{li2019citywide, gu2020exploiting}, and enhance traffic efficiency through congestion management~\cite{wang2016traffic, ji2022stden}.

% \IEEEPARstart{T} he growth of the urban population has presented huge challenges to urban transportation and sustainability. 
% % Because of the potential to 
% {\cheng To}
% tackle these challenges, Intelligent Transportation Systems (ITS) has become a trending topic in both academia and industry recently~\cite{ji2022stden}. As the core technology of ITS, traffic prediction across the whole city is crucial for many important urban management scenarios~\cite{zhang2020spatial}. For example, accurate traffic flow prediction can not only mitigate tragedies caused by the sudden traffic flow spike to maintain public safety, but also enable timely effective traffic controls to improve transportation efficiency. In general, traffic prediction aims to forecast the future traffic volume from past traffic observations. 

Existing studies on ST data usually take a temporal and spatial perspective. The temporal perspective describes the data evolution trends over time, which 
% are mainly divided into 
covers patterns of
closeness, periodicity, and trend~\cite{zhang2017deep, guo2019attention}. Researchers apply the recurrent neural networks (\eg LSTM)~\cite{yang2019urban, li2018dcrnn_traffic}, temporal convolutional networks~\cite{wu2019graph, wang2022traffic}, or attention mechanism~\cite{yao2019revisiting} to encode the temporal features of traffic series. On the other hand, the spatial perspective mainly describes the relations of different spatial regions or sensors in the study area~\cite{ji2020interpretable, bai2020adaptive}. All the relations are mixed together to form a global graph structure. Main efforts deploy graph neural networks for holistic spatial dependency modeling~\cite{yu2018spatio, guo2021learning, ji2023spatial}, or attention mechanism for spatial information aggregation across the whole graph~\cite{pan2022spatio, wang2022traffic}. 

Despite their notable successes, the existing methods generally adopt a holistic scheme, \ie characterizing ST data on the original graph structure as a whole. They ignore the nuances among partial ST data affected by different latent factors. Each factor may only affect relevant graph nodes and relations rather than the whole. 
Taking traffic ST data as an example, the formation of traffic flow typically follows the human mobility law in real-world socioeconomic interactions~\cite{alessandretti2020scales}. It is driven by many complex latent factors, such as commuting, entertainment, business, \etc. Each of them focuses on different urban regions and forms different traffic flows~\cite{wang2019understanding, wang2014discovering, fan2014cityspectrum, wang2017community}. 
The complex and entangled relations among multiple latent factors bring an urge for decoupling these factors in ST data prediction. This remains unexplored by the existing works. As a result, the predicted ST data contains a mixture of entangled factors, inevitably harming interpretability and causing difficulties in thoroughly understanding the evolution of ST data.

In this paper, we propose a \textit{multi-factor spatio-temporal prediction} task to tackle the multi-factor issue. The task aims to forecast the ST data evolution of multiple latent factors separately and then integrate all partial results for the final ST prediction. 
% Although the traditional traffic prediction problem has been studied broadly in many papers, multi-factor traffic prediction faces the following two unique challenges. $i)$ \textit{Effective theoretical solution} for multi-factor traffic prediction. The solution should consider the existence of multiple factors and be theoretically proven to be effective in reducing prediction error. In this way, its instantiation can successfully model the impacts of multiple factors on traffic status. $ii)$ \textit{Tailored portable framework} designed for multi-factor traffic prediction. Since factor-dependent traffic flows are not available in real scenarios, multi-factor traffic prediction can only utilize external knowledge. This implies that the framework should be well-designed by the principles extracted from human knowledge. Moreover, the framework should be portable to existing graph-based traffic prediction models since the theoretical solution behind it is generic. 
%
%
% To tackle these challenges, we raise 
We design a novel \underline{S}patio-\underline{T}emporal \underline{G}raph \underline{D}ecomposition \underline{L}earning framework (\model) capable of the multi-factor spatio-temporal prediction. Specifically, we \textbf{first} propose a theoretical solution called the decomposed prediction strategy. Following the divide-and-conquer paradigm, the strategy decomposes the original graph structure that mixes multiple factors into several subgraphs. Then, we can learn relevant ST data on each subgraph separately and integrate them to produce an overall traffic prediction. More importantly, we show that the proposed strategy is theoretically guaranteed to 
% reduce the prediction error. 
have a smaller prediction error than directly predicting on the original graph.
\textbf{Next}, we instantiate the strategy as a framework called \model, whose key components are automatic graph decomposition and decomposed learning networks. The former component is implemented by matrix masking with two regularization terms preserving the principles of graph completeness and subgraph independence. The latter component is based on forward and backward residual connections and a stack of ST blocks. Each block is responsible for learning ST data on the subgraph relevant to a certain factor. We conduct extensive experiments on four public ST benchmarks including two grid graph-based datasets and two network graph-based datasets. The results show that our \model can enhance the performance of a wide range of graph-based ST models by an average of 9.41\%. Our contributions are four-fold:
% \\
% \bitem To our knowledge, this research is pioneering in terms of the multi-factor issue in ST prediction. In light of this, we propose a multi-factor ST prediction task and a theoretically guaranteed portable framework (\model) capable of solving the task.\\  
% \bitem We propose a decomposed prediction strategy to solve the multi-factor prediction problem via graph decomposition in a divide-and-conquer fashion. We also provide a theoretical analysis of the strategy's ability to reduce prediction errors.\\
% \bitem We design a portable framework called \model to instantiate the proposed strategy by an automatic graph decomposition and a decomposed learning network. Our \model can learn disentangled traffic embedding for each latent factor.\\ 
% \bitem Extensive experiments are conducted on four real-world traffic datasets, demonstrating the performance gain achieved by the \model plugin on a variety of graph-based ST models. Moreover, a case study confirms that learned patterns can provide interpretability for deep models and their predictions.
\begin{itemize}[leftmargin=*]
    \item To our knowledge, this research is pioneering in dealing with the multi-factor issue for ST prediction. In light of this, we introduce a multi-factor ST prediction task and successfully solve it with a theoretically guaranteed portable framework.  
    \item We propose a decomposed prediction strategy to achieve multi-factor ST prediction via graph decomposition in a divide-and-conquer fashion. We prove that the strategy can reduce ST prediction errors.
    % We provide a theoretical analysis of the strategy's ability to reduce prediction errors.
    \item We design a portable \model framework to instantiate the proposed strategy by automatic graph decomposition and decomposed learning networks. The framework can learn disentangled ST embedding for each latent factor. 
    \item Extensive experiments on four real-world ST datasets exhibit the performance gain achieved by the \model plugin on various graph-based ST models. 
    Moreover, a case study confirms that learned patterns can provide interpretability for deep models and their predictions.
\end{itemize}

The remainder of the paper is organized as follows: Sec.~\ref{sec:pre} introduces the basic concepts and the problem of multi-factor ST prediction. Sec.~\ref{sec:theory} proposes a theoretical solution for this problem, while Sec.~\ref{sec:method} instantiates it as an ST graph
decomposition learning framework. Sec.~\ref{sec:expt} evaluates the proposed framework using four public datasets with various baselines. The related work and conclusion are presented in Sec.~\ref{sec:related_work} and \ref{sec:con}, respectively.

\section{Preliminaries}\label{sec:pre}

This section gives the basic concepts and introduces the problem definition.

\begin{mydef}[Spatio-Temporal Graph]
A spatio-temporal (ST) graph is defined as $\mathcal{G}^{(t)} = (G, \bm{X}^{(t)})$. The \textbf{graph structure} is represented as $G = (\mathcal{V}, \mathcal{E}, \bm{A})$, where $\mathcal{V}$ is a set of nodes with the size of $|\mathcal{V}| = N$, and $\mathcal{E}$ is a set of edges connecting two nodes in $\mathcal{V}$. 
% The node can be a region, a sensor, \etc. 
$\bm A \in \mathbb{R}^{N \times N}$ denotes the adjacent matrix of graph $G$. The \textbf{graph signal} $\bm{X}^{(t)} \in \mathbb{R}^{N \times F}$ denotes the ST observations defined on $G$ at the $t$-th time slot, where $F$ is the feature channel.
% $F$ is the number of flow types of each node (\eg inflow, outflow).
\end{mydef}

\noindent \textbf{Problem Statement}. Let $\mathcal{G}^{(t-T:t)} = (G, \bm{X}^{(t-T:t)})$ be the past ST data with time window $T$. The key of most ST prediction methods, including ours, is to derive an ST model $f(\cdot)$ to forecast the future evolution of ST data, \ie $\bm{X}^{(t+1)}= f(\mathcal{G}^{(t-T:t)})$. In this work, we reformulate the problem as \textit{multi-factor spatio-temporal prediction}. It aims to learn a multi-factor ST model $f(\cdot) = \{f_k(\cdot)\}_{k=1}^{K}$, where $K$ is the number of latent factors. Consequently, $\bm{X}^{(t+1)}$ is expected to consist of $K$ independent components, \ie $\bm{X}^{(t+1)} = \sum_{k=1}^{K} \bm{X}_{k}^{(t+1)}$, where $\bm{X}_{k}^{(t+1)} = f_k(\mathcal{G}^{(t-T:t)})$. The $k$-th component $\bm{X}_{k}^{(t+1)}$ is for characterizing the aspect of $\mathcal{G}^{(t-T:t)}$ that is relevant to the $k$-th latent factor.

\section{Decomposed Prediction Strategy}\label{sec:theory}

In this section, we first provide a theoretical solution for the multi-factor ST prediction problem via a divide-and-conquer paradigm. It is called the decomposed prediction strategy. Then, we prove the effectiveness of the proposed strategy.

\subsection{Strategy Framework}

\paratitle{ST Data Generation Perspective.}
From the data generation perspective, the generation process of ST data can be affected by multiple latent factors related to natural phenomena or human socioeconomic activities. 
% As shown in \fig~\ref{fig:data_gen}, 
Assuming there are $K$ factors, the ST graph data $\mathcal{G}$ can be regarded as a mixture of all $\mathcal{G}_k = (G_k, \bm{X}_k), k\in[1, K]$. We use $\mathcal{G}_k$ to denote the ST graph affected by the $k$-th latent factor. This effect involves two aspects: graph structure $G$ and graph signal $\bm{X}$. Intuitively, we should decouple the graph structures corresponding to different factors, and then model the graph signal under the impact of each factor separately. 

% In contrast, directly modeling the traffic data with a mixture of multiple factors will bring about the stacking of uncertainties from different factors, which reduces the predictability of traffic data. This will make the traffic prediction problem more difficult and thus degrade the model performance.

\paratitle{Steps of Decomposed Prediction Strategy.}
Since our decomposed prediction strategy is in the framework of the divide-and-conquer approach, there are three steps in our proposed strategy as in \tab~\ref{tab:analogy}. Details are in the following.
\begin{itemize}[leftmargin=*]
    \item \textbf{Graph decomposition}. The ST graph structure $G = (\mathcal{V}, \mathcal{E}, \bm{A})$, containing a mixture of different latent factors, is decomposed into several subgraphs $\{G_k\}_{k=1}^{K}$. The subgraph $G_k = (\mathcal{V}_k, \mathcal{E}_k, \bm{A}_k)$ characterizes the $k$-th latent factor. There are two constraints in the decomposition process: graph completeness $\bigcup \mathcal{E}_k = \mathcal{E}$, and subgraph independence $\bigcap \mathcal{E}_k = \emptyset$. 
    \item \textbf{Decomposed learning}. We learn the relevant ST graph signal data $\bm{X}_k$ over $G_k$ separately by a factor-specific ST encoder $f_k(\cdot)$. 
    \item \textbf{Prediction integration}. We integrate the partial prediction results corresponding to each latent factor and obtain the overall ST prediction via $\bm{X} = \sum_{k=1}^{K}\bm{X}_k$.
\end{itemize}
% 三步: 1.图分解, 2. 解耦学习, 3. 合并预测

\paratitle{Advantage of Decomposed Prediction Strategy.}
The proposed strategy can reduce the difficulty of the ST prediction problem, and thus produce smaller errors. Specifically, directly modeling the ST data generated by multiple complex factors will result in the stacking of uncertainties from different factors, which reduces the \textit{predictability} of ST data~\cite{zhang2022beyond, song2010limits, wang2015predictability}. This will make the ST prediction problem more difficult and thus degrade the model performance. In contrast, inspired by the divide-and-conquer approach, the decomposed prediction strategy can characterize the impact of each latent factor separately. It makes the partial ST data $\bm{X}_k$ over the decomposed $k$-th subgraph easier to model than the data over the original graph structure, \ie $\bm{X}$. In this way, we obtain an easier problem, namely \textit{multi-factor ST prediction}. The advantage of the proposed strategy is summarized in the following theorem.

\begin{theorem}\label{thm:stgy}
Assume the error lower bound of an ST prediction problem is $E_o$. After introducing the decomposed prediction strategy and transforming the original problem into a multi-factor ST prediction problem, we can have an error lower bound $E_d$ with $E_d < E_o$.
\end{theorem}

In \thm~\ref{thm:stgy}, a smaller error lower bound means the difficulty of the corresponding problem is lower, while a low-difficulty problem indicates that it can be solved with smaller errors. The proof of \thm~\ref{thm:stgy} is provided in the following part. 

\subsection{Theoretical Analysis}

% \subsubsection{Proof of Strategy Advantage}

We here prove the advantage of our decomposed prediction strategy given in \thm~\ref{thm:stgy}. As shown in \tab~\ref{tab:analogy}, our strategy includes three steps related to three parts of the divide-and-conquer approach. Therefore, we first transform these three parts into three following lemmas and then prove \thm~\ref{thm:stgy}.
% with their aid.

\begin{table}[t]
    \centering
    \caption{Analogy from the decomposed prediction strategy to divide-and-conquer approach.}\label{tab:analogy}%\vspace{-.1cm}
    \begin{tabular}{|rcl|}
    \hline 
    Graph decomposition & $\longrightarrow$ & Divide \\
    Decomposed learning  & $\longrightarrow$ & Conquer \\
    Prediction integration & $\longrightarrow$ & Combine \\
    \hline
    \end{tabular}
\end{table}

\begin{lemma}[Divide]\label{lem:divide}
    The graph decomposition step divides the original ST prediction problem into a number of subproblems that are smaller and independent instances of the original problem. 
\end{lemma}

\begin{lemma}[Conquer]\label{lem:conquer}
    The decomposed learning step conquers the subproblems by solving them separately. The error lower bound of the $k$-th subproblem is $2e_k \sigma_k^2$. $e_k$ is the lower bound of error rate, and $\sigma_k^2$ is the variance of the ST random variable $X_k$ corresponding to the $k$-th subproblem.
\end{lemma}

\begin{lemma}[Combine]\label{lem:combine}
    The prediction integration step combines the results of the subproblems into the solution for the original problem. The error lower bound of the combined solution is $E_d = \sum_{k=1}^{K} 2 e_k \sigma_k^2$.
\end{lemma}

The proof of the three lemmas can be found in \apx~\ref{appx:lemmas}.
% A of the supplementary materials.
% \apx~\ref{appx:proof}.
% We now delve into the proof of \thm~\ref{thm:stgy} by combing the conclusions of Lemma~\ref{lem:divide}, Lemma~\ref{lem:conquer}, and Lemma~\ref{lem:combine}.
Before we delve into the proof of \thm~\ref{thm:stgy}, we have to emphasize that the error rate lower bound of the original ST prediction problem $e$ satisfies
\begin{equation}\label{eq:err_rate}
    \forall k \in [1, K], e_k < e,
\end{equation}
where $e_k$ is the error rate lower bound of the $k$-th subproblem, which defines the minimal probability of making incorrect predictions. \equ~\eqref{eq:err_rate} is proved in \apx~\ref{appx:predictability} by using information entropy and predictability~\cite{cover2006elements, brabazon2008natural}. Next, we give the proof of \thm~\ref{thm:stgy}.

\begin{IEEEproof}
Assume the error lower bound of the original ST prediction problem $P$ is $E_o$. From Lemma~\ref{lem:conquer}, we know that $E_o$ can be defined as
% From Lemma~\ref{lem:conquer}, we know that $E_o$ is proportional to the error rate lower bound and the variance of the corresponding traffic random variable. It can be defined as 
% Similar to \equ~\eqref{eq:elbo_k_ana}, we can derive the error lower bound of the original problem as
\begin{equation}\label{eq:elbo_org}
E_o = 2 e \sigma^2, ~~\st~~ \sigma^2 = \sum_{k=1}^{K} \sigma_k^2
\end{equation}
where $e$ is the error rate lower bound of directly solving the original ST prediction problem. $\sigma^2$ is the variance of the ST random variable $X$ corresponding to the original problem $P$. The constraint $\sigma^2 = \sum_{k=1}^{K} \sigma_k^2$ is derived from Lemma~\ref{lem:divide}. Because the original problem is divided into $K$ independent subproblems, the covariance of any two subproblems is zero. This indicates variances of all ST random variables $X_k$ sum up to the variance of $X$.

Comparing $E_d$ in Lemma~\ref{lem:combine} and $E_o$ in \equ~\eqref{eq:elbo_org} with the help of \equ~\eqref{eq:err_rate}, we have
\begin{equation}
    E_d = \sum_{k=1}^{K} 2 e_k \sigma_k^2 < \sum_{k=1}^{K} 2 e \sigma_k^2 = 2 e \sigma^2 = E_o.
\end{equation}
This means that after introducing the decomposed prediction strategy, we can have a smaller error lower bound compared with directly solving the original problem.

\end{IEEEproof}

\section{Instantiation Framework}\label{sec:method}

\begin{figure}[t]
    \centering
    \includegraphics[width=1.0\columnwidth]{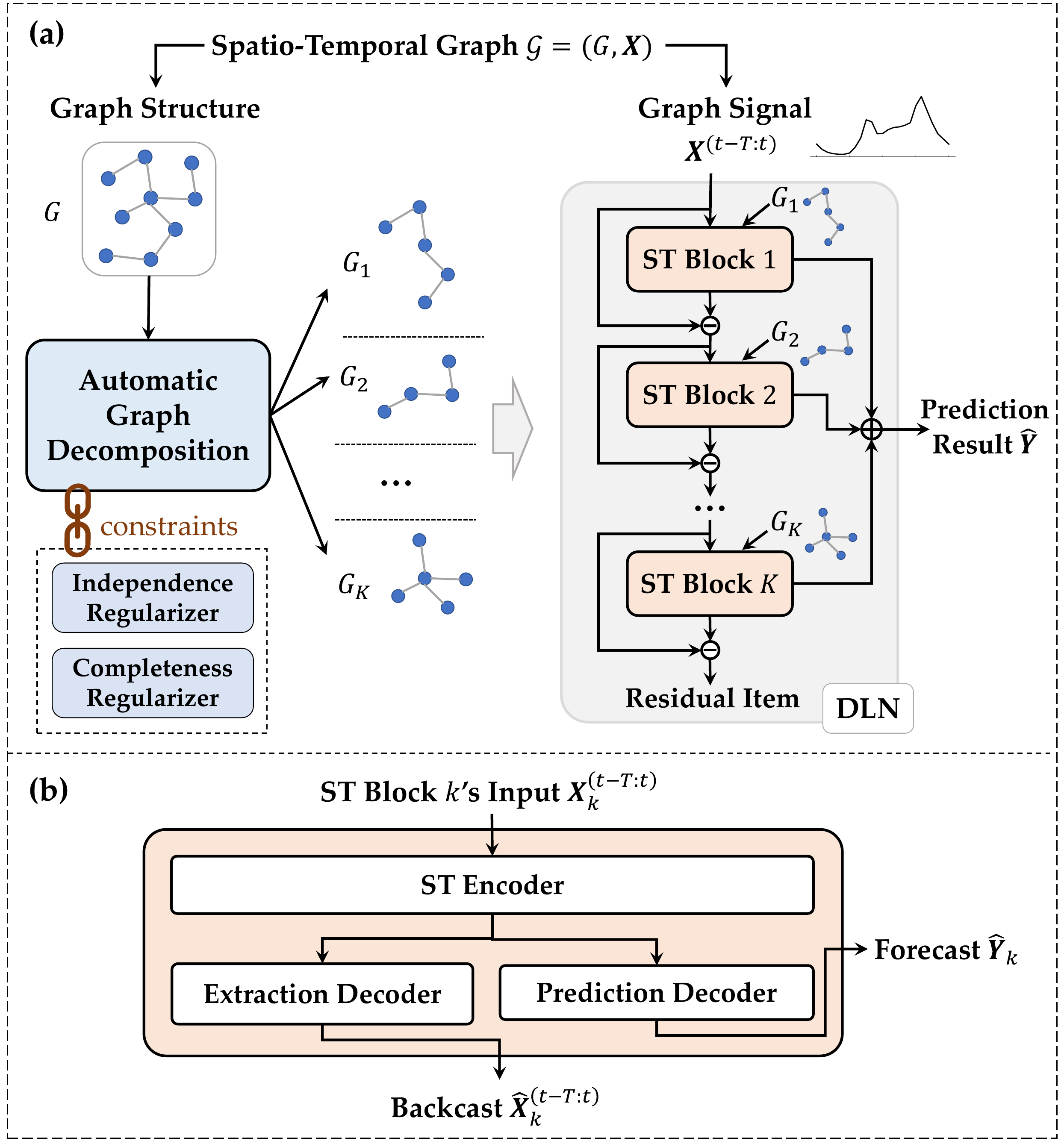}\vspace{-.4cm}
    \caption{(a) The architecture of our \model framework. The graph structure of the input ST graph is decomposed as several subgraphs via Automatic Graph Decomposition constrained by completeness and independence regularizers. Then, ST graph signals are fed into the Decomposed Learning Network (DLN) for ST prediction corresponding to each subgraph and the overall graph. The residual item of DLN is expected to be zero after subtracting recovered all subgraphs' signals from the input graph signals. (b) An illustration of the ST Block.}
    \label{fig:framework}
\end{figure}

Following the principles of the decomposed prediction strategy in Sec.~\ref{sec:theory}, we instantiate a portable Spatio-Temporal Graph Decomposition Learning (\model) framework as shown in \fig~\ref{fig:framework}. In this section, we will elaborate on the \model with technical details.

\subsection{Overview Framework}

Classical data-driven methods model ST data directly based on the original ST graph structure $G$ that is affected by multiple latent factors. Differently, guided by the decomposed prediction strategy, we propose \model to model the influence of each factor separately. 
As illustrated in \fig~\ref{fig:framework}, our \model consists of two main components: $i)$ an Automatic Graph Decomposition (AGD) component that corresponds to the first step of the decomposed prediction strategy, and $ii)$ a Decomposed Learning Network (DLN) that corresponds to the remaining two steps of the strategy. In AGD, we propose a matrix masking-based automatic method to decompose the original graph into multiple subgraphs.  
Each subgraph can characterize the graph structure relevant to a certain factor.
Details are given in Sec.~\ref{ssec:agd}.

% \begin{figure}[t]
%     \centering
%     \includegraphics[width=1.0\columnwidth]{figures/framework.pdf}
%     \caption{The architecture of our \model framework. The graph structure of the input ST graph is decomposed as several subgraphs via Automatic Graph Decomposition constrained by completeness and independence regularizers. Then, ST graph signals are fed into the Decomposed Learning Network (DLN) for ST prediction corresponding to each subgraph and the overall graph. The residual item of DLN is expected to be zero after subtracting recovered all subgraphs' signals from the input graph signals. An illustration of the ST Block is provided in \fig~\ref{fig:stb}.}
%     \label{fig:framework}
% \end{figure}

% \begin{figure}[t]
%     \centering
%     \subfigure[The architecture of our \model framework. The graph structure of the input ST graph is decomposed as several subgraphs via Automatic Graph Decomposition constrained by completeness and independence regularizers. Then, ST graph signals are fed into the Decomposed Learning Network (DLN) for ST prediction corresponding to each subgraph and the overall graph. The residual item of DLN is expected to be zero after subtracting recovered all subgraphs' signals from the input graph signals.]{\includegraphics[width=1.0\columnwidth]{figures/framework.pdf}}
%     \subfigure[he architecture of the spatio-temporal block.]{\includegraphics[width=0.75\columnwidth]{figures/stb.pdf}}    
%     \caption{Illustarions of the \model framework and ST block.}
%     \label{fig:framework}
% \end{figure}

Over these subgraphs, we design a deep learning model DLN. It stacks several ST blocks by a dual residual mechanism that consists of subtractive and additive residual connections. The former is to disentangle ST data relevant to each ST block, which is relevant to the second step of our strategy. The latter aims to combine partial predictions of every ST block to produce the overall ST prediction. Details of DLN are provided in Sec.~\ref{ssec:dln}.

% In each block, we $i)$ employ an ST encoder to preserve the ST contextual information into ST embeddings, and $ii)$ introduce two different decoders to forecast future ST data evolution of the relevant subgraph and extract historical ST observations of the same subgraph simultaneously. 

In our \model, the graph decomposition process is designed to be learnable. Therefore, it can be unified into the same framework with the decomposed learning network. In this way, the whole framework can be easily trained in an end-to-end manner.

\subsection{Automatic Graph Decomposition}\label{ssec:agd}

This part proposes a simple yet efficient graph decomposition scheme. It is implemented by matrix masking and incorporated with two regularization terms preserving the principles of graph completeness and subgraph independence. 

\subsubsection{Matrix Masking-based Decomposition}

Given an original graph structure $G = \left({\mathcal{V}}, {\mathcal{E}}, {\bm A}\right)$ of ST graph data, this module aims to decompose it into a set of subgraphs $\{G_k\}_{k=1}^{K}$. Specifically, for the $k$-th subgraph $G_k = \left({\mathcal{V}}_k, {\mathcal{E}}_k, {\bm A}_k\right)$, we generate the corresponding adjacency matrix $\tilde{\bm A}_k$ as follows: 
\begin{equation}
    \tilde{\bm A}_k = \psi_1(\bm{M}_k) \bm A,
\end{equation}
where the masking matrix $\bm M_k \in \mathbb{R}^{N\times N}$ is learnable. The function $\psi_1(\cdot) = (\mathrm{Tanh}(\cdot) + 1)/2$ generate an edge ratio in the range $(0, 1)$. It ensures that the edge weights in $\tilde{\bm A}_k$ do not exceed that in $\bm A$. When $\bm A$ is a binary adjacency matrix, each item in $\tilde{\bm A}_k$, \ie $\tilde{a}_{k, i, j}$, can denote the probability that the relevant edge exists. Compared with heuristic graph decomposition methods~\cite{xie2014distributed, ng2001spectral, ji2020interpretable, wang2022traffic}, our approach is simple, differentiable, and has no restrictions on the type of input graphs.

\subsubsection{Regularization Terms}

There are two constraints in the graph decomposition step of the decomposed prediction strategy: $i)$ the graph completeness $\bigcup \mathcal{E}_k = \mathcal{E}$, and $ii)$ the subgraph independence $\bigcap \mathcal{E}_k = \emptyset$. To satisfy both constraints, we introduce two corresponding regularization terms.
% including a completeness regularizer and an independence regularizer. 

Specifically, the \textit{completeness regularizer} $\mathcal{L}_{c}$ aims to minimize the difference between the original graph and the reconstructed one. It is defined by the graph adjacency matrix as follows:
\begin{equation}\label{eq:loss_c}\small
    \mathcal{L}_{c} = \left\Vert \psi_2(\bm{A}) - \psi_2({\hat{\bm{A}} })\right\Vert_1, ~\st~ \hat{\bm{A}} = \left(\sum_{k=1}^{K} \bm{A}_k\right),  
\end{equation}
where $\Vert \cdot \Vert_1$ is the $\ell_1$ norm that first flattens the input matrix. $\hat{\bm{A}}$ is the reconstructed adjacency matrix of the original graph $\bm{A}$. We use $\psi_2(\cdot)$ to normalize $\hat{A}$ as a binary matrix. Specifically, we implement it by $\psi_2(x) = (\mathrm{Tanh}(4(x - 0.5)) + 1)/2$. 

The \textit{independence regularizer} $\mathcal{L}_i$ aims to maintain the independence between different subgraphs. It is defined as:
\begin{equation}\label{eq:loss_i}\small
    \mathcal{L}_i = \frac{1}{K(K-1)}\sum_{k=1}^{K} \sum_{j=1, j\neq k}^{K} \left\Vert\bm{A}_k^\top \bm{A}_j\right\Vert_1.
\end{equation}
Minimizing $\mathcal{L}_i$ can steer any pair of adjacency matrices to be orthogonal, thereby making the corresponding subgraphs independent.

\subsection{Decomposed Learning Network}\label{ssec:dln}

The right part of \fig~\ref{fig:framework} presents the architecture of the decomposed learning network (DLN). It is comprised of $K$ spatio-temporal blocks and a dual residual mechanism. $K$ is the number of decomposed subgraphs in Sec.~\ref{ssec:agd}. We briefly overview the framework as follows.
\begin{itemize}[leftmargin=*]
    \item The \textit{ST block} aims to learn ST data related to the corresponding subgraph. It takes residual historical ST observations of the previous block as input and produces two outputs: predicted ST data and extracted historical ST data. Then, the predicted data of all blocks are integrated for the final forecast, while the extracted historical data are used for backward estimation, \aka backcast. 
    \item The \textit{dual residual mechanism} has two branches: one runs on the backcast of each ST block and the other runs on the overall traffic forecast of all blocks. The former uses subtractive residual connections and the latter is compromised of additive residual connections.
\end{itemize}

In the following parts, we will illustrate the details of these two components.

% \begin{figure}[t]
%     \centering
%     \includegraphics[width=0.75\columnwidth]{figures/stb.pdf}
%     \caption{The architecture of the spatio-temporal block, \ie ST block.}
%     \label{fig:stb}
% \end{figure}

\subsubsection{Spatio-Temporal Block}

As shown in \fig~\ref{fig:framework}(b), the ST block aims to learn ST data relevant to different subgraphs via backcast and forecast. Formally, the inputs of the $k$-th ST block are $\bm{X}_k^{(t-T:t)} \in \mathbb{R}^{T \times N \times F}$ and the adjacency matrix of $k$-th subgraph, $\bm{A}_k$. They are mapped by the ST blocks into two outputs:
\begin{equation}\small
    \left\{\hat{\bm{X}}_k, \hat{\bm{Y}}_k\right\} = \mathrm{STB}\left(\bm{X}_k^{(t-T:t)}, \bm{A}_k\right),
\end{equation}
where $\hat{\bm{X}}_k^{(t-T:t)}$ is the estimation of $\bm{X}_k^{(t-T:t)}$, and $\hat{\bm{Y}}_k \in \mathbb{R}^{N \times 2}$ is the ST data prediction of time step $t+1$, whose ground truth is $\bm{X}_k^{(t+1)}$.
For the very first ST block in DLN, its respective input $\bm{X}_k^{(t-T:t)}$ is the overall model input, \ie the historical ST data with a lookback window $T$. For the rest of the ST blocks, their inputs $\bm{X}_k^{(t-T:t)}$ are residual outputs of the previous ST blocks. 

% In each block, we $i)$ employ an ST encoder to preserve the ST contextual information into ST embeddings, and $ii)$ introduce two different decoders to forecast future ST data evolution of the relevant subgraph and extract historical ST observations of the same subgraph simultaneously. 
Internally, the ST block consists of two parts. The first part is an ST encoder that produces ST embeddings using block inputs. The second part consists of a prediction decoder and an extraction decoder. It uses the ST embeddings to produce the forecast output $\hat{\bm{Y}}_k$ and the backcast output $\hat{\bm{X}}_k^{(t-T:t)}$. Many deep ST models were proposed to encode ST data as ST embeddings and then make predictions using the embeddings~\cite{yu2018spatio, wu2019graph, ji2023spatial}. Therefore, we can directly adopt the encoding subnetworks of these models as our ST encoder, and adopt their forecasting subnetworks as our prediction decoder. As for the extraction decoder, since the main difference between the prediction decoder and the extraction decoder is the output length, we can get the extraction decoder by simply modifying the output length of the prediction decoder.

\subsubsection{Dual Residual Mechanism}

We introduce a dual residual mechanism to disentangle the partial ST data affected by different latent factors. It consists of two types of residual connections: subtractive and additive residual connections. 

The subtractive residual connection removes components of ST block input that are not helpful for ST prediction of the downstream blocks. For the $k$-th ST block, the subtractive residual connection is defined by
\begin{equation}\small
    \bm{X}_k^{(t-T:t)} = \bm{X}_{k-1}^{(t-T:t)} - \hat{\bm{X}}_{k-1}^{(t-T:t)},
\end{equation}
where $\hat{\bm{X}}_{k-1}^{(t-T:t)}$ is the backcast of the input of $(k-1)$-th ST block, \ie $\bm{X}_{k-1}^{(t-T:t)}$. The backcast residual branch $\bm{X}_k^{(t-T:t)}$ can be considered as running a disentanglement process of the input ST data. The previous block removes the portion of ST data $\hat{\bm{X}}_{k-1}^{(t-T:t)}$ defined on $(k-1)$-th subgraph, making the prediction task of the downstream blocks irrelevant to $(k-1)$-th latent factor. In this way, we encourage the ST block to disentangle ST data affected by different factors. To ensure that no additional ST data is left behind, we design a residual loss for the last residual item to be close to zero. It is defined as
\begin{equation}\small
    \mathcal{L}_{r} = \left\Vert \bm{X}_K^{(t-T:t)}\right\Vert_1,
\end{equation}
where $\Vert \cdot \Vert_1$ is the $\ell_1$ norm that first flattens the input data.

On the other hand, the additive residual connection aggregates the partial ST predictions of every ST block to produce the overall ST prediction. It can be described by
\begin{equation}\label{eq:res_add}\small
    \hat{\bm{Y}} = \sum_{k=1}^{K} \hat{\bm{Y}}_k.
\end{equation}
Since this allows for the $k$-th ST block to focus on the partial ST data affected by the $k$-th latent factors, we expect accurate predictions of ST data that mix impacts of several latent factors. \equ~\eqref{eq:res_add} also provides a disentanglement of future ST data because all partial predictions can be treated as the decomposition of $\hat{\bm{Y}}$.  
% It is noted that the dual residual mechanism also facilitates more fluid gradient backpropagation.
We finally optimize the multi-factor ST prediction task by minimizing the loss function:
\begin{equation}\label{eq:loss_p}\small
    \mathcal{L}_p = \left\Vert\hat{\bm{Y}} - \bm{X}^{(t+1)} \right\Vert_1.
\end{equation}
% where $\Vert \cdot \Vert_1$ is the $\ell_1$ norm that first flattens the input data.

More importantly, with the guidance of expert knowledge, the prediction of each ST block can have its own physical meaning in reality. This can enhance the interpretability of our model's outputs. For example, in ST traffic prediction, the first ST block could be responsible for commuting flow, the second for entertaining flow, and so on.

% More importantly, with proper guidance of human mobility knowledge, the predicted flow of each ST block can have its own physical meaning in reality, such as the commuting flow and entertaining flow, which enhance the interpretability of the model's outputs. 

\subsection{Model Training}

In the learning process of our \model, we calculate the overall loss by incorporating the graph decomposition losses in \equ~\eqref{eq:loss_c} and \equ~\eqref{eq:loss_i} and the decomposed learning loss in \equ~\eqref{eq:loss_p} into the joint learning objection:
\begin{equation}\small
    \mathcal{L}_{joint} = \mathcal{L}_c + \mathcal{L}_i + \mathcal{L}_r + \mathcal{L}_p.
\end{equation}

Our model can be trained end-to-end via the back-propagation algorithm. The entire training procedure can be summarized into Algo.~\ref{alg:train}. In lines 1-4, we construct training data. In lines 6-12, we iteratively optimize \model by gradient descent until the stopping criterion is met. Specifically, in lines 7-8, we first select a random batch of data and then apply the forward-backward operation on the whole model to get gradients of all parameters. At last, in lines 9-12, we update the parameters within AGD and DLN components by gradient descent respectively. 

\begin{algorithm}[t]\small
    \caption{Training Algorithm of \model}\label{alg:train}
    \begin{algorithmic}[1] %[1] enables line numbers
        \Require{Traffic flow graph data $\{\mathcal{G}^{(1)}, \dots, \mathcal{G}^{(\tau)}\}$.}
        \Ensure{The learned \model model.}
        \For{\textit{available} $t \in \{1, \dots, \tau\}$}\label{alg:data_s}
        \State $\mathrm{X} \leftarrow \{\bm{X}^{(t-T:t)}, \bm{A}\}$. \Comment{Input data}
        \State $\mathrm{Y} \leftarrow \bm{X}^{t+1}$.\Comment{Label}
        \State Put $\{\mathrm{X}, \mathrm{Y}\}$ into $\mathcal{D}_{train}$.
        \EndFor\label{alg:data_e}
        \State Initialize all trainable parameters.
        \While{\textit{stopping criterion is not met}}\label{alg:train_s}
        \State Randomly select a batch $\mathcal{D}_{batch}$ from $\mathcal{D}_{train}$.
        \State Forward-backward on $\mathcal{L}_{joint}$ by $\mathcal{D}_{batch}$.
        \For{\textit{parameter} $\theta$ \textit{in AGD}}\Comment{AGD is in Sec.~\ref{ssec:agd}}
        \State $\theta = \theta - \alpha \cdot  \nabla_{\theta}\mathcal{L}_{joint}$\Comment{$\alpha$ is learning rate}
        \EndFor
        \For{\textit{parameter} $\theta$ \textit{in DLN}}\Comment{DLN is in Sec.~\ref{ssec:dln}}
        \State $\theta = \theta - \alpha \cdot \nabla_{\theta}\mathcal{L}_{joint}$
        \EndFor
        \EndWhile\label{alg:train_e}
        \State \textbf{return} Learned \model model.
    \end{algorithmic}
\end{algorithm}

% \subsection{Discussion on Portability}
% Basic deep models: FNN, GRU, transformer
% Static Graph-based ST models: STGCN, STSSL
% Static-Dynamic Hybrid Graph-based ST models: GWNet, MTGNN

\section{Experiments}\label{sec:expt}

In this section, we evaluate the effectiveness of \model on a series of experiments over four real-world benchmark datasets, which are summarized to answer the following research questions:

\begin{itemize}[leftmargin=*]
    \item \textbf{RQ1}: How is the overall traffic prediction performance of \model as compared to various baselines?
    \item \textbf{RQ2}: How effective is the decomposition strategy when ported to various baseline methods?
    \item \textbf{RQ3}: How do the designed different submodules contribute to the model performance?
    \item \textbf{RQ4}: What is the impact of hyper-parameters in \model?
    \item \textbf{RQ5}: How do the learned representations of decomposed graphs contribute to model interpretability?
\end{itemize}

\subsection{Experimental Settings}

Since urban traffic data is the most typical kind of spatio-temporal data, we test our model mainly based on traffic data and the ST traffic prediction task. 

\subsubsection{Data Description} 
We evaluate our model on two types of graph-based public real-world traffic datasets, including grid graph-based datasets and network graph-based datasets. 
The grid graph-based datasets include a bike dataset and a taxi dataset. Bike data contains bike rental records. Taxi data record the number of taxis coming to and departing from a region given a specific time interval.
The network graph-based datasets include two sensor datasets. They are mainly generated by highway sensors that measure the flow volume or the vehicle speed of given roads.
These datasets are generated by millions of taxis, bikes, or highway vehicles on average and contain thousands of time steps and hundreds of regions or roads. The statistical information is summarized in \tableautorefname~\ref{tab:dataset}. More details are in the following:

\begin{itemize}[leftmargin=*]
    \item \textbf{NYCBike} is a public traffic flow dataset collected from New York City. Specifically, NYCBike contains bike rental data over a period of six months from 04/01/2014 to 09/30/2014~\cite{zhang2017deep}. We divide New York City into a raster with $16\times 8$ grid zones and map bike records as traffic flow among zones. Traffic flows in NYCBike are measured on an hourly basis, and the total sequence length is 4,392. 
    \item \textbf{NYCTaxi} is also a public dataset collected from New York City. Specifically, NYCTaxi contains taxi GPS ranging from 01/01/2015 to 03/01/2015. The city is divided into a grid of $20 \times 10$ to map taxi trajectories as traffic flow among zones~\cite{yao2019revisiting}. It is measured every 30 minutes, and the total sequence length is 2,880.
    \item \textbf{PEMSD7(M)} is a public traffic dataset collected from California Performance of Transportation (PeMS)~\cite{chen2001freeway}. Specifically, it contains data from 228 sensors in District 07 over a period of 2 months from 05/01/2012 to 06/30/2012~\cite{yu2018spatio}. The traffic information is recorded every 5 minutes. We aggregate them on a 30-minute basis and obtain 2,112 sequences.
    % a the total number of time slices is 12,672.
    \item \textbf{PEMSD8} is a public traffic dataset collected from PeMS. Specifically, it contains data from 170 sensors in District 08 over a period of 2 months from 07/01/2018 to 08/31/2018~\cite{guo2019attention}. The traffic information is originally recorded every 5 minutes. We aggregate them in 30 minutes and obtain 2,976 sequences.
    % the total number of time slices is 17,856.
\end{itemize}

For all datasets, previous 2-hour flows and previous 3-day flows around the predicted time are used to predict the flows for the next time step. This can facilitate the modeling of shifted temporal correlations~\cite{yao2019revisiting}. 
A sliding window strategy is utilized to generate samples. Then, we split each dataset into the training, validation, and test sets with a ratio of 7:1:2. 

\paratitle{Construction of ST Graph Structure.} 
For grid graph-based datasets (\ie NYCBike and NYCTaxi), to distinguish the weights of edges between different nodes, we compute the dynamic time warping (DTW) distance of traffic series between nodes as the adjacency matrix. Specifically, the element $a_{ij}$ of the adjacency matrix $\bm{A}$ is computed by $a_{ij}=~~\mathrm{DTW}(\bm{x}_i, \bm{x}_j)$, where $\bm{x}_i$ denotes the time series for node $i$, and $r$ is a pre-defined threshold that determines the sparsity of the resulting adjacency matrix.
For network graph-based datasets (\ie PEMSD7(M) and PEMSD8), we compute the pairwise road network distances between sensors and build the adjacency matrix using thresholded Gaussian kernel~\cite{li2018dcrnn_traffic}.

\begin{table}[t]\small
    \centering
    \caption{Statistics of all four datasets.}
    \setlength{\tabcolsep}{1.5mm}
    \begin{tabular}{r|cc|cc}
      \toprule 
      {Data type} & \multicolumn{2}{c|}{Grid graph} & \multicolumn{2}{c}{Network graph} \\
      \midrule
      Dataset & NYCBike  & NYCTaxi & PeMSD7(M) & PeMSD8 \\
      Time interval & 1 hour & 30 min & 30min & 30min \\
      \# graph nodes & 128 & 200 & 228 & 170 \\
      seq. length & 4,392 & 2,880 & 2,112 & 2,976 \\
      % \# vehicle & 6.8k+ & 22m+ & TODO & TODO\\
      \bottomrule
    \end{tabular}%
    \vspace{.1cm}
    \label{tab:dataset}%
  \end{table}%

\begin{table*}[ht]
    \centering
    \caption{ST prediction on two grid graph-based traffic datasets, \ie NYCBike and NYCTaxi, in terms of MAE, MAPE (\%), and RMSE. In and Out represents the inflow and outflow. ``Improv.'' measures the improvement of \model over the counterpart baseline with respect to each task and each metric.}
    \resizebox{\linewidth}{!}{
    \setlength{\tabcolsep}{1mm}
    \renewcommand{\arraystretch}{1.5}
      \begin{tabular}{|c||cc|cc|cc||cc|cc|cc|}
      \hline
        \textbf{Dataset}    & \multicolumn{6}{c||}{\textbf{NYCBike}} & \multicolumn{6}{c|}{\textbf{NYCTaxi}} \\
      \hline
        Metric & \multicolumn{2}{c|}{MAE}& \multicolumn{2}{c|}{MAPE} & \multicolumn{2}{c||}{RMSE} & \multicolumn{2}{c|}{MAE}& \multicolumn{2}{c|}{MAPE} & \multicolumn{2}{c|}{RMSE}\\
      \hline
        Type & In & Out& In & Out& In & Out& In & Out& In & Out& In & Out\\
      \hline
        HA & 10.85 & 11.09 & 45.68 & 46.98 & 16.91 & 17.50 & 36.76 & 28.51 & 46.72 & 44.02 & 66.87 & 56.68 \\
        %ARIMA & 10.66 & 11.33 & 33.05 & 35.03 & . & . & 20.86 & 16.80 & 21.49 & 21.23 & . & .\\
        SVR & 8.48 & 8.59 & 34.63 & 34.63 & 13.67 & 14.31 & 35.81 & 28.65 & 45.62 & 43.47 & 68.91 & 58.05 \\
      \hline
STGCN & 5.26$\pm$0.02 & 5.53$\pm$0.04 & 22.58$\pm$0.29 & 23.13$\pm$0.38 & 7.65$\pm$0.05 & 8.23$\pm$0.11 & 12.87$\pm$0.21 & 10.93$\pm$0.13 & 17.13$\pm$0.16 & 17.23$\pm$0.2 & 23.09$\pm$0.5 & 23.62$\pm$0.47\\
STGCN+ & 5.07$\pm$0.05 & 5.36$\pm$0.02 & 21.77$\pm$0.08 & 22.55$\pm$0.05 & 7.36$\pm$0.11 & 7.95$\pm$0.07 & 11.84$\pm$0.12 & 9.68$\pm$0.15 & 17.14$\pm$0.84 & 16.80$\pm$0.01 & 21.17$\pm$0.26 & 17.61$\pm$0.79\\
\rowcolor{gray!20}Improv. & +3.61\%&+3.07\%&+3.59\%&+2.51\%&+3.79\%&+3.40\%&+8.00\%&+11.44\%&-0.06\%&+2.50\%&+8.32\%&+25.45\%\\
   \hline

% \hline
GWNet & 5.15$\pm$0.02 & 5.43$\pm$0.02 & 22.67$\pm$0.21 & 23.24$\pm$0.20 & 7.38$\pm$0.03 & 8.01$\pm$0.04 & 
12.7$\pm$0.02 & 10.39$\pm$0.08 & 17.45$\pm$1.12 & 17.79$\pm$0.48 & 22.48$\pm$0.06 & 18.29$\pm$0.12\\
GWNet+ & 4.97$\pm$0.02 & 5.25$\pm$0.01 & 21.61$\pm$0.21 & 22.28$\pm$0.02 & 7.12$\pm$0.04 & 7.75$\pm$0.05 & 
11.53$\pm$0.08 & 9.32$\pm$0.04 & 15.78$\pm$0.21 & 16.28$\pm$0.32 & 20.57$\pm$0.21 & 16.15$\pm$0.07\\
\rowcolor{gray!20}Improv.&+3.50\%&+3.31\%&+4.68\%&+4.13\%&+3.52\%&+3.25\%&+9.21\%&+10.30\%&+9.57\%&+8.49\%&+8.50\%&+11.70\%\\
      \hline
MTGNN & 5.26$\pm$0.03 & 5.57$\pm$0.03 & 22.77$\pm$0.10 & 23.83$\pm$0.61 & 7.68$\pm$0.06 & 8.31$\pm$0.03 & 12.71$\pm$0.03 & 10.19$\pm$0.03 & 17.03$\pm$0.06 & 17.71$\pm$0.25 & 22.37$\pm$0.05 & 18.0$\pm$0.12\\
MTGNN+ & 5.12$\pm$0.01 & 5.42$\pm$0.01 & 22.28$\pm$0.13 & 23.18$\pm$0.03 & 7.46$\pm$0.04 & 8.12$\pm$0.02 & 11.94$\pm$0.15 & 9.71$\pm$0.09 & 16.38$\pm$0.36 & 16.98$\pm$0.47 & 21.07$\pm$0.26 & 16.9$\pm$0.19\\
\rowcolor{gray!20}Improv.&+2.66\%&+2.69\%&+2.15\%&+2.73\%&+2.86\%&+2.29\%&+6.06\%&+4.71\%&+3.82\%&+4.12\%&+5.81\%&+6.11\%\\

      \hline
MSDR & 5.65$\pm$0.10 & 6.00$\pm$0.14 & 24.54$\pm$0.59 & 25.89$\pm$0.10 & 8.26$\pm$0.15 & 8.99$\pm$0.23 & 18.49$\pm$1.20 & 15.89$\pm$1.02 & 25.27$\pm$1.73 & 25.78$\pm$1.93 & 33.7$\pm$2.09 & 33.17$\pm$1.67\\
MSDR+ & 5.25$\pm$0.05 & 5.55$\pm$0.07 & 23.22$\pm$0.24 & 23.86$\pm$0.30 & 7.68$\pm$0.11 & 8.31$\pm$0.10 & 14.72$\pm$0.10 & 11.53$\pm$0.28 & 20.61$\pm$1.00 & 20.37$\pm$0.64 & 26.31$\pm$0.34 & 21.44$\pm$0.63\\
\rowcolor{gray!20}Improv.&+7.08\%&+7.50\%&+5.38\%&+7.84\%&+7.02\%&+7.56\%&+20.39\%&+27.44\%&+18.44\%&+20.99\%&+21.93\%&+35.36\%\\
      \hline
STSSL & 5.18$\pm$0.03 & 5.44$\pm$0.02 & 22.37$\pm$0.17 & 23.00$\pm$0.21 & 7.56$\pm$0.06 & 8.15$\pm$0.03 & 12.19$\pm$0.14 & 10.0$\pm$0.09 & 16.43$\pm$0.04 & 17.06$\pm$0.17 & 21.7$\pm$0.28 & 19.05$\pm$0.37\\
STSSL+ & 5.10$\pm$0.04 & 5.37$\pm$0.04 & 22.36$\pm$0.10 & 22.99$\pm$0.26 & 7.45$\pm$0.14 & 8.03$\pm$0.16 & 11.83$\pm$0.10 & 9.74$\pm$0.14 & 16.36$\pm$0.01 & 17.11$\pm$0.30 & 20.86$\pm$0.22 & 17.06$\pm$0.43\\
\rowcolor{gray!20} Improv.&+1.54\%&+1.29\%&+0.04\%&+0.04\%&+1.46\%&+1.47\%&+2.95\%&+2.60\%&+0.43\%&-0.29\%&+3.87\%&+10.45\%\\
    \hline
    \end{tabular}
    }%
    \label{tab:overall}%
\end{table*}

\subsubsection{Evaluation Metrics \& Baselines}

% \paratitle{Evaluation Metrics.} 
We adopt three commonly used metrics to measure the accuracy of the prediction results, including Mean Average Error (MAE), Mean Absolute Percentage Error (MAPE), and Rooted Mean Square Error(RMSE). 
% They can be expressed as:
% \begin{equation}
% \begin{aligned}
%     \mathrm{MAE} &=\frac{1}{M}\sum_{i=1}^{M} \left\Vert\bm{Y}_{i} - \hat{\bm{Y}}_{i}\right\Vert_{1},\\
%     \mathrm{MAPE} &=\frac{1}{M} \sum_{i=1}^{M} \left\Vert\frac{\bm{Y}_{i} - \hat{\bm{Y}}_{i}}{\bm{Y}_{i}}\right\Vert_{1},\\
%     \mathrm{RMSE} &=\frac{1}{M}\sum_{i=1}^{M} \left\Vert\bm{Y}_{i} - \hat{\bm{Y}}_{i}\right\Vert_{2}.
% \end{aligned}
% \end{equation}
% where $M$ indicates the number of test samples, $\bm{Y}_{i} \in \mathbb{R}^{N \times F}$ represent the $i$-th ground truth, and $\hat{\bm{Y}}_{i}$ is the predicted result. $\Vert \cdot \Vert_{1}$ and $\Vert \cdot \Vert_{2}$ denotes the L1 and L2 norm that first flattens the input data.
% \paratitle{Baselines.}
As for the baselines, we first compare our \model with two classic non-deep models, including:
\begin{itemize}[leftmargin=*]
    \item \textbf{HA}: Historical Average models ST data as a seasonal process and calculates the average of data as prediction results.
    \item \textbf{SVR}: Support Vector Regression is a regression model using a linear support machine. It is a traditional machine learning method.
\end{itemize}

Secondly, we implement five deep learning models proposed in recent years to compare with our \model.

\begin{itemize}[leftmargin=*]
\item{\textbf{STGCN}}~\cite{yu2018spatio}: Spatio-Temporal Graph Convolution Network is a graph convolution-based model that uses 1D convolution to capture spatial dependencies and temporal correlations.
\item{\textbf{GWNet}}~\cite{wu2019graph}: GWNet is short for graph-wavenet that combines diffusion graph convolution with dilated causal convolution to capture spatial-temporal dependencies.
\item{\textbf{MTGNN}}~\cite{wu2020connecting}: Multivariate Time series forecasting with Graph Neural Networks employs adaptive graphs and integrates GRU with graph convolutions.
\item{\textbf{MSDR}}~\cite{liu2022msdr} Multi-step Dependency Relation Networks is an RNN-based model that explicitly takes hidden states of multiple historical steps as the input of each time unit.
\item{\textbf{STSSL}}~\cite{ji2023spatial}: Spatio-Temporal Self-Supervised Learning is a self-supervised model that designs two auxiliary tasks to learn effective region embeddings for traffic prediction.
\end{itemize}

For the deep learning baselines, we implement them based on the released codes of the LibCity~\cite{wang2021libcity} benchmark~\footnote{\url{https://github.com/LibCity/Bigscity-LibCity}}, which integrates a great number of ST forecasting models according to their original paper. Note that all these deep models can be considered as combinations of ST encoders and ST decoders. They can be used as backbone models and integrated with our \model framework. Next, we give the settings of \model in detail.

\subsubsection{Parameter Settings}

The \model is implemented with Python 3.9 and PyTorch 1.12.0. The number of subgraphs is searched from the range of $\{4, 5, 6, 7, 8, 9, 10\}$ for all datasets. We found the optimal settings are $6, 6, 6, 8$ for NYCBike, NYCTaxi, PEMSD7(M), and PEMSD8. The batch size is set to 32. The model training phase is performed using the Adam optimizer with the learning rate set as 0.001. The hidden dimension is determined by the different backbone models. For any more details, readers could refer to our public code repository~\footnote{\code}.

\subsection{Performance Comparison (RQ1 \& RQ2)}

In this part, we report and analyze the traffic forecasting performance of non-deep models, base deep models, and enhanced deep models over grid graph-based datasets (\tab~\ref{tab:overall}) and network graph-based datasets (\tab~\ref{tab:overall_2}). For simplicity, we name the deep models enhanced by our \model as ``base model+''. We ran all deep models and their enhanced versions with five different random seeds, and then reported the average performance and its standard deviation, shown as ``mean$\pm$standard deviation''.

\begin{table*}[ht]%\small
\begin{threeparttable}
    \centering
    \caption{Traffic forecasting on two network graph-based datasets, \ie PEMSD7(M) and PEMSD8, in terms of MAE, MAPE (\%), and RMSE.}
    % \resizebox{\linewidth}{!}{
    \setlength{\tabcolsep}{1mm}
    \renewcommand{\arraystretch}{1.5}
      \begin{tabular}{|c||ccc||ccc|}
      \hline
        \textbf{Dataset}    & \multicolumn{3}{c||}{\textbf{PEMSD7(M)}} & \multicolumn{3}{c|}{\textbf{PEMSD8}} \\
      \hline
        Metric & MAE& MAPE & RMSE& MAE& MAPE & RMSE\\
      \hline
        HA & 4.62 & 11.77 & 7.50 & 36.89 & 24.66 & 53.99 \\
        %ARIMA & . & . & . & . & . & . \\
        SVR & 3.23 & 8.72 & 6.14 & 30.90 & 18.39 & 47.10 \\
      \hline
STGCN   &2.15$\pm$0.01&5.14$\pm$0.07&4.25$\pm$0.03&11.56$\pm$0.25&8.25$\pm$0.13&19.55$\pm$0.60\\
%STGCN+  &2.04$\pm$0.01&4.88$\pm$0.01&4.16$\pm$0.01&8.70$\pm$0.03&6.72$\pm$0.05&14.67$\pm$0.07\\
STGCN+&2.04$\pm$0.01(+5.12\%)\tnote{1}&4.88$\pm$0.01(+5.06\%)&4.16$\pm$0.01(+2.12\%)&8.70$\pm$0.03(+24.74\%)&6.72$\pm$0.05(+18.55\%)&14.67$\pm$0.07(+24.96\%)\\

\hline
GWNet      &2.13$\pm$0.01&5.01$\pm$0.02&4.22$\pm$0.02&9.63$\pm$0.06&7.31$\pm$0.12&15.97$\pm$0.11\\
%GWNet+     &2.04$\pm$0.01&4.82$\pm$0.04&4.17$\pm$0.01&8.34$\pm$0.06&6.24$\pm$0.27&14.38$\pm$0.07\\
GWNet+&2.04$\pm$0.01(+4.23\%)&4.82$\pm$0.04(+3.79\%)&4.17$\pm$0.01(+1.18\%)&8.34$\pm$0.06(+13.4\%)&6.24$\pm$0.27(+14.64\%)&14.38$\pm$0.07(+9.96\%)\\
\hline
MTGNN   &2.06$\pm$0.01&4.82$\pm$0.01&4.15$\pm$0.02&8.93$\pm$0.07&7.54$\pm$0.50&15.17$\pm$0.06\\
%MTGNN+  &2.04$\pm$0.02&4.78$\pm$0.05&4.15$\pm$0.01&8.27$\pm$0.03&6.46$\pm$0.13&14.54$\pm$0.07\\
MTGNN+&2.04$\pm$0.02(+0.97\%)&4.78$\pm$0.05(+0.83\%)&4.15$\pm$0.01(+0.00\%)&8.27$\pm$0.03(+7.39\%)&6.46$\pm$0.13(+14.32\%)&14.54$\pm$0.07(+4.15\%)\\
\hline
MSDR    &3.40$\pm$0.07&8.49$\pm$0.29&6.19$\pm$0.22&17.78$\pm$1.10&15.38$\pm$1.78&28.03$\pm$1.64\\
%MSDR+   &2.53$\pm$0.19&6.19$\pm$0.57&5.00$\pm$0.40&14.45$\pm$0.66&12.20$\pm$0.98&23.43$\pm$1.45\\
MSDR+&2.53$\pm$0.19(+25.59\%)&6.19$\pm$0.57(+27.09\%)&5.00$\pm$0.40(+19.22\%)&14.45$\pm$0.66(+18.73\%)&12.20$\pm$0.98(+20.68\%)&23.43$\pm$1.45(+16.41\%)\\
\hline
STSSL   &2.09$\pm$0.02&4.98$\pm$0.03&4.19$\pm$0.02&10.77$\pm$0.02&7.97$\pm$0.20&18.01$\pm$0.08\\
%STSSL+  &2.05$\pm$0.01&4.87$\pm$0.03&4.20$\pm$0.01&8.02$\pm$0.05&6.10$\pm$0.18&14.22$\pm$0.04\\

STSSL+&2.05$\pm$0.01(+1.91\%)&4.87$\pm$0.03(+2.21\%)&4.20$\pm$0.01(-0.24\%)&8.02$\pm$0.05(+25.53\%)&6.10$\pm$0.18(+23.46\%)&14.22$\pm$0.04(+21.04\%)\\

    \hline
    \end{tabular}
    % }%
    \label{tab:overall_2}%
    \begin{tablenotes}
       \item [1] The percentage measures the improvement of \model over the counterpart baseline with respect to one metric, \eg MAE.
     \end{tablenotes}
\end{threeparttable}
\end{table*}

\subsubsection{Results on Grid Graph-Based Datasets}

\tab~\ref{tab:overall} shows the comparison results on grid graph-based datasets, including NYCBike and NYCTaxi. 

\paratitle{Performance Superiority of \model.}
We first compare the enhanced deep models by our \model with their basic version. On average, the enhanced models have 3.63\% MAE improvement, 3.31\% MAPE improvement, and 3.66\% RMSE improvement in NYCBike, as well as 10.31\% MAE improvement, 6.80\% MAPE improvement, and 13.75\% RMSE improvement in NYCTaxi. The enhanced model GWNet+ performs the best among all methods. It is worth noting that the overall improvement in the NYCTaxi dataset is about twice that of the NYCBike dataset. The reason for this phenomenon is that there are more traffic participants in NYCTaxi (22m+) compared to NYCBike (6.8k+). This produces traffic patterns affected by more complex factors in NYCTaxi, compared with NYCBike. In this case, \model learns the factor-specific patterns by a decomposed prediction strategy, enabling higher performance gains.  
In the case of NYCBike, where traffic patterns are simple, the original model can achieve good performance using powerful deep learning techniques, so the performance improvement of \model is insignificant. As a result, our \model provides a greater performance boost to deep models on NYCTaxi compared to NYCBike.

\paratitle{Performance Comparison between Baselines.}
We can observe that $i$) Conventional non-deep models, including ARIMA and SVR, are not good enough to predict traffic flows, because of the limitation of the model expressiveness that is unable to capture complex and dynamic ST correlations. $ii$) Compared with non-deep models, deep models deliver better forecasting performance, because they exploit a wide variety of deep learning techniques to target the learning of high-quality ST representations.

\subsubsection{Results on Network Graph-Based Datasets}

\tab~\ref{tab:overall} shows the comparison results on network graph-based datasets, including PEMSD7(M) and PEMSD8.

\paratitle{Performance Superiority of \model.}
We first compare the enhanced deep models by our \model with their basic version. On average, the enhanced models have 7.56\% MAE improvement, 7.80\% MAPE improvement, and 4.46\% RMSE improvement in the PEMD7(M) dataset, as well as 17.96\% MAE improvement, 18.33\% MAPE improvement, and 15.31\% RMSE improvement in the PEMSD8 dataset. Compared to grid graph-based datasets, the results on network graph-based datasets are more promising. Notably, when applying our \model on MTGNN, the best base model, it can still significantly improve the MAE (7.39\%), MAPE (14.32\%), and RMSE (4.15\%) in the PEMSD8 dataset. The reason behind this is that PEMSD8 has 170 sensors corresponding to many highway roads, and they may be responsible for different functions. Therefore, a model collectively learning mixed data evolution patterns cannot effectively capture such diverse ST traffic trends. Instead, our \model employs a decomposed learning network to learn different evolution trends individually, enabling the model to better capture the mixed ST patterns. As a result, the deep models enhanced by \model can have much better performance than the basic versions.

\paratitle{Performance Comparison between Baselines.}
First, non-deep models still produce unsatisfactory ST traffic prediction performance on network graph-based datasets, showing their limitations for modeling complex real-world ST data. Second, with the boost of advanced deep learning techniques, deep models generate better results compared to non-deep ones. 
Interestingly, for the base version, the performance ranking of the best three models is MTGNN $>$ GWNet $>$ STSSL, but the ranking shows as STSSL+ $>$ GWNet+ $>$ MTGNN+ when enhanced by \model. This suggests that \model has different enhancement effects for different types of deep models.

\subsection{Ablation Study (RQ3)}

To verify the effectiveness of submodules in \model, we conduct ablation studies on all four datasets regarding three important submodules. They include automatic graph decomposition, regularization terms for decomposition, and the decomposed learning network. There are five models enhanced by our \model, we only report the results of STGCN in this part because other models exhibit a similar phenomenon. For clarity, we use \model to denote the enhanced deep models.

\begin{figure}[t]
    \centering
    \includegraphics[width=\columnwidth]{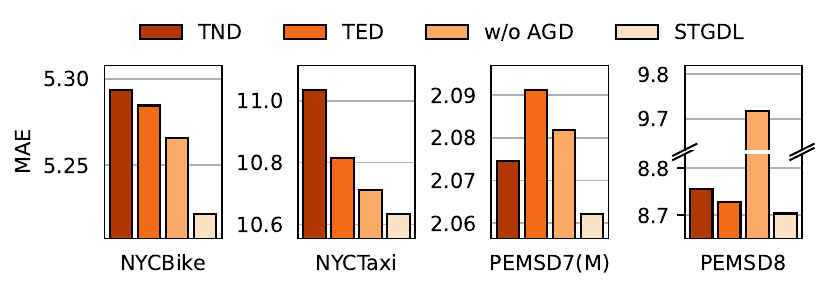}
    \vspace{-.7cm}
    \caption{Effectiveness of the automatic graph decomposition \wrt MAE.}
    \label{fig:ablation-2}
\end{figure}

\subsubsection{Effectiveness of Automatic Graph Decomposition}

Before we delve into the ablation study on automatic graph decomposition, we first introduce some background of graph decomposition. Graph decomposition method, \aka graph partitioning, can be categorized into node decomposition method and edge decomposition method according to the way of splitting the input graph data~\cite{bulucc2016recent, feldmann2015balanced}. The node decomposition method aims to assign the graph nodes to each subgraph and maintain edges between nodes of the same subgraph~\cite{ng2001spectral}. This may cause edges between different subgraphs to be cut off, thereby destroying the original graph structure. Differently, the edge decomposition method assigns the edges of the input graph to each subgraph, and each group of assigned edges constitutes a subgraph~\cite{xie2014distributed}, which may lead to the redundancy of boundary nodes but maintain the original graph structure. 

In Sec.~\ref{ssec:agd}, we introduced an Automatic Graph Decomposition (AGD) method, which belongs to the category of edge decomposition. In order to verify its effectiveness, we designed three variants using traditional decomposition methods or even removing the decomposition module. We describe their details in the following.

\begin{itemize}[leftmargin=*]
    \item \textbf{TED}: A Traditional Edge Decomposition (TED) algorithm is employed to generate the subgraphs for each ST block. Here we choose a classic method called degree-based hashing~\cite{xie2014distributed} which assigns each edge by hashing the identification of its end-vertex with a lower degree.
    \item \textbf{TND}: A Traditional Node Decomposition (TND) algorithm is used for subgraph generation for each ST block. We choose spectral clustering~\cite{ng2001spectral} which groups graph nodes using eigendecomposition of the Laplacian matrix of the input graph.
    \item \textbf{w/o AGD}: We remove the AGD submodule, and we use the original graph as the input graph for each ST block.
\end{itemize}

The results are presented in \fig~\ref{fig:ablation-2}. Primarily, we can observe that the performance ranking is TND $<$ TED $<$ \model in most cases, which shows the superiority of the automatic method. Because the ST data vary with time, the static decomposition approach cannot effectively capture the dynamic correlation between different graph nodes, thereby delivering unsatisfactory results. In contrast, our \model can capture such dynamics via the data-driven automatic decomposition method and produces good performance. On the other hand, TND is beaten by TED in most cases. This illustrates the importance of maintaining a complete graph structure in ST prediction, hence we chose the edge decomposition paradigm over the node decomposition paradigm. Moreover, w/o AGD always performs worse than \model. This indicates the necessity of decomposition, without which a model is unable to decompose the ST traffic data relevant to multiple latent factors. Instead, it directly performs mixed-factor ST prediction that is complicated, leading to worse performance compared with \model. By combining the automatic fashion and edge decomposition method, our \model produces the best performance.

\subsubsection{Effectiveness of Regularization Terms}

\begin{figure}[t]
    \centering
    \includegraphics[width=\columnwidth]{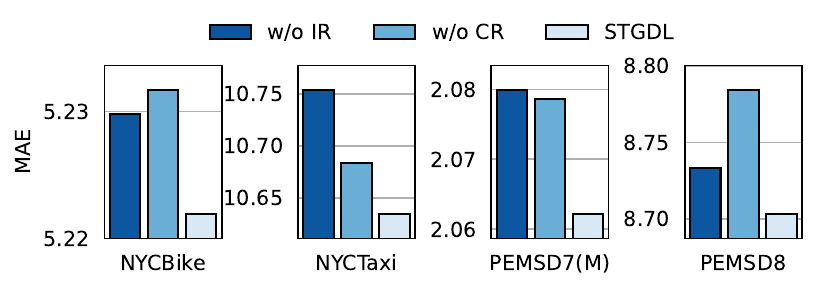}
    \vspace{-.7cm}
    \caption{Effectiveness of regularization terms \wrt MAE.}
    \label{fig:ablation-3}
\end{figure}

There are two constraints within the decomposed prediction strategy, including graph completeness and subgraph independence. To this end, we design two regularization terms for automatic graph decomposition, \ie completeness regularizer $\mathcal{L}_{c}$ in \equ~\eqref{eq:loss_c} and independence regularizer $\mathcal{L}_i$ in \equ~\eqref{eq:loss_i}, to satisfy these constraints. To examine their effectiveness, we prepared two variants as follows.

\begin{itemize}[leftmargin=*]
    \item \textbf{w/o IR}: The independence regularizer $\mathcal{L}_i$ is removed.
    \item \textbf{w/o CR}: The completeness regularizer $\mathcal{L}_c$ is removed.
\end{itemize}

The results are shown in \fig~\ref{fig:ablation-3}. It can be observed that across all four datasets, the predictive accuracy of the two variants without regularization terms is inferior to that of the original model. 
This verifies the significant role of both regularization terms.
Specifically, the objective of $\mathcal{L}_i$ is maximizing the sum of dot products between adjacency matrices of different subgraphs. This ensures the distinctiveness between subgraphs and orthogonality among the subproblems relevant to each subgraph, thereby enforcing different ST blocks to focus on different data evolution patterns.
Removing $\mathcal{L}_{i}$ would result in the non-orthogonality of the subproblems, violating the orthogonality constraint and leading to performance degradation.
On the other hand, the completeness regularizer $\mathcal{L}_{c}$ aims to ensure that every edge in the original graph is present in some decomposed subgraph. Eliminating $\mathcal{L}_{c}$ would cause certain edges from the original graph to be absent in any subgraphs, resulting in information loss and, consequently, a decline in ST prediction performance.

\subsubsection{Effectiveness of Decomposed Learning Network}

In the decomposed learning network (Sec.~\ref{ssec:dln}), we design a dual residual mechanism to disentangle ST data. The mechanism includes additive and subtractive residual connections among ST blocks. The additive ones are designed for integrating partial ST predictions of each block for the overall result. The subtractive ones are designed for removing data patterns that have been learned by the backcast process of the previous ST block. These patterns do not contribute to the ST prediction of downstream blocks. 
To verify the effectiveness of this mechanism, we designed three variants to compare with our \model in the following.

\begin{itemize}[leftmargin=*]
    \item \textbf{w/o BC}: The backcast part in the ST block is removed, and we use the original input data as the input for each ST block.
    \item \textbf{w/o SC}: The subtractive connections are removed. We feed the output of the previous ST block as the input of the current ST block.
    \item \textbf{w/o SAC}: The subtractive and additive connections are both removed, which means the dual residual mechanism is dropped. This variant simply stacks several ST blocks.
\end{itemize}

\begin{figure}[t]
    \centering
    \includegraphics[width=0.9\columnwidth]{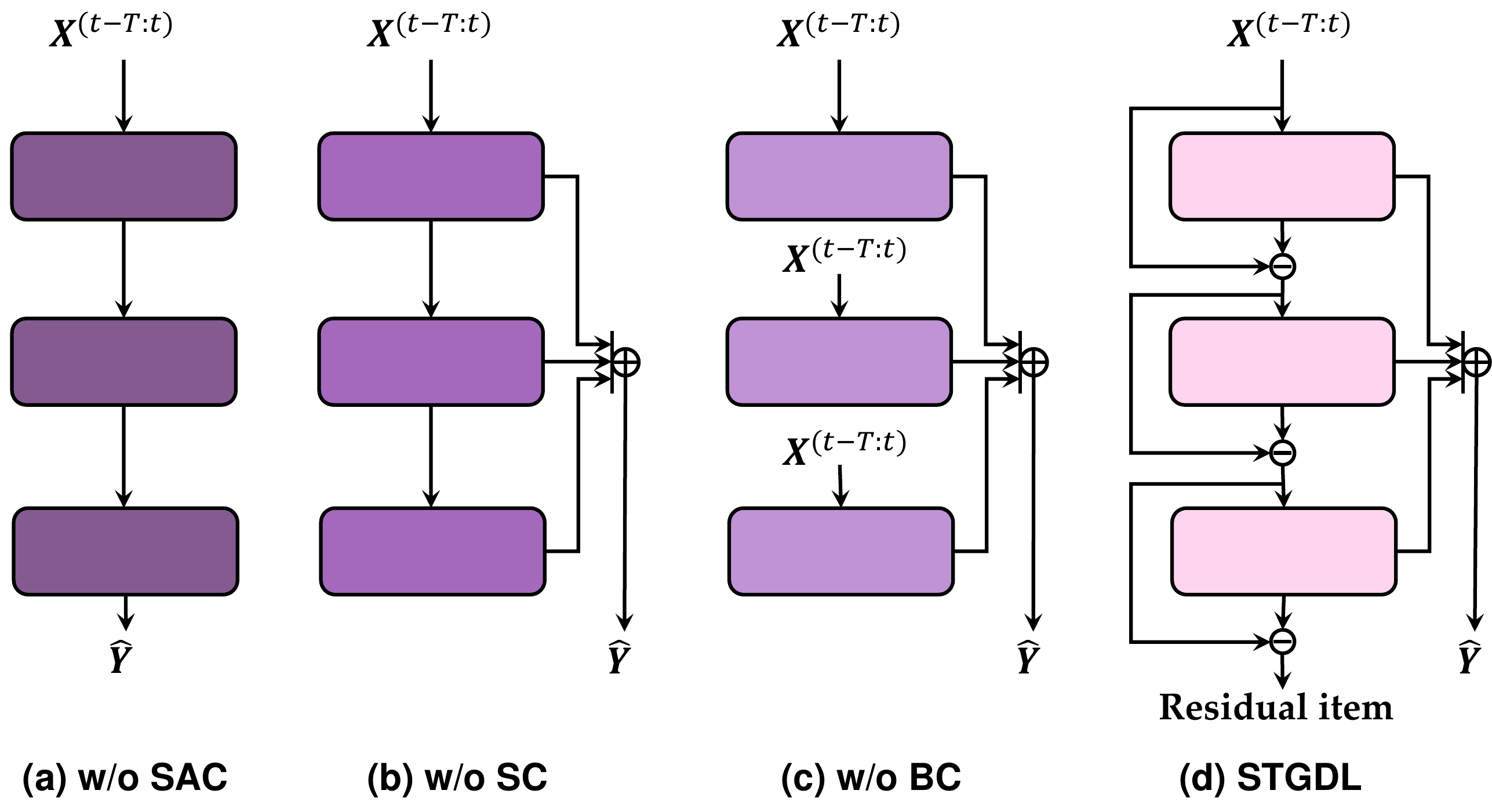}
    \vspace{-.2cm}
    \caption{Illustration of variants about the dual residual mechanism. Without loss of generality, we take three ST blocks as an example. There can be more blocks in practice.}
    \label{fig:sac}
\end{figure}

\begin{figure}[t]
    \centering
    \includegraphics[width=\columnwidth]{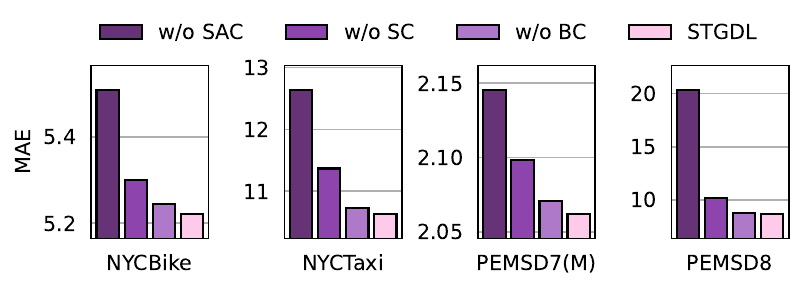}
    \vspace{-.5cm}
    \caption{Effectiveness of the decomposed learning network \wrt MAE.}
    \label{fig:ablation-4}
\end{figure}

To facilitate understanding, we illustrate all three variants and \model in \fig~\ref{fig:sac}. 
The comparison results are depicted in \fig~\ref{fig:ablation-4}. 
First, \model consistently outperforms w/o SAC with a large margin, indicating the effectiveness of our dual residual mechanism. Since ST data are usually affected by different latent factors, without this mechanism, w/o SAC struggles to make accurate predictions for traffic ST data that mix the impact of several factors. 
It is worth noting that the dual residual mechanism not only solves the problem of multi-factor ST data modeling, but also brings \textit{better model scalability}. This allows ordinary ST models to go deeper, thus preparing for large ST models.

Second, the performance of other variants, \ie w/o BC and w/o SC that have modified the data propagation process, is lower than that of the original model. This verifies the importance of designing a proper data flow in model architecture. Recall the subtractive connections are responsible for eliminating data components affected by the current block-specific factor. Since this portion of data is irrelevant for downstream blocks, subtractive connections can facilitate downstream ST prediction. However, w/o BC feeds all ST blocks with the original input data, forcing each ST block to extract relevant data components from the complex and mixed original data distribution. In contrast, only the first ST block needs to do that in our \model, and the extraction complexity of the remaining ST blocks is linearly decreasing. This facilitates the learning efficiency.
As for the variant w/o SC, it replaces the subtractive connections with the regular connections, so the model must learn both the extraction operation and the subtraction operation simultaneously. The results in \fig~\ref{fig:ablation-4} show that this severely burdens the model learning and makes the prediction accuracy drop significantly.

\subsection{Parameter Sensitivity (RQ4)}

\begin{figure}[t]
    \centering
    \includegraphics[width=0.95\columnwidth]{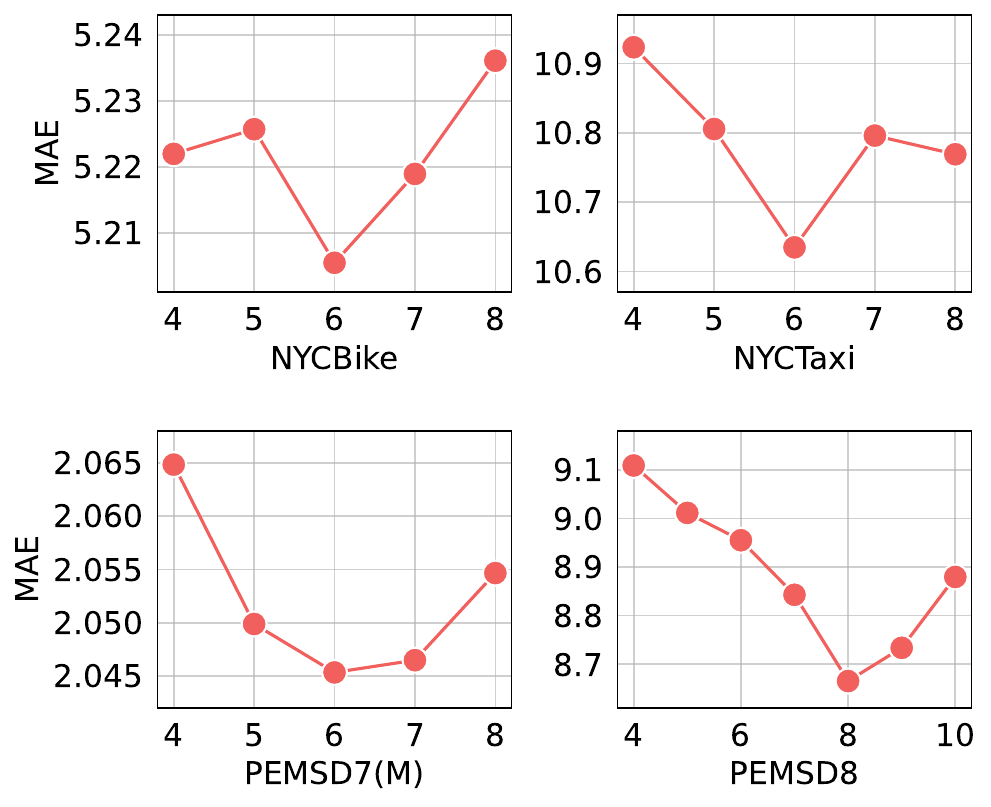}
    \vspace{-.4cm}
    \caption{Evaluation on the number of subgraphs $K$ \wrt MAE. The optimal choice of $K$ for NYCBike, NYCTaxi, PEMSD7(M), and PEMSD8 is 6, 6, 6, and 8, respectively.}
    \label{fig:hyper}\vspace{-.4cm}
\end{figure}

In this part, we conduct experiments to analyze the impact of the only hyper-parameter in \model, \ie subgraph number $K$.
We vary $K$ in the range of $\{4, 5, 6, 7, 8, 9, 10\}$ for all four datasets. We use MAE as the default metric and STGCN as the default base model. The results are shown in \fig~\ref{fig:hyper}. The optimal choice of $K$ is 6, 6, 6, and 8 for NYCBike, NYCTaxi, PEMSD7(M), and PEMSD8 respectively. As we can see from the results, having too few subgraphs can seriously affect the performance because it fails to capture the effects of multiple factors. Too many subgraphs may lead to overfitting, which reduces the prediction accuracy. When applying $K=9, 10$ for the first three datasets, the performance drops significantly so we omit them in \fig~\ref{fig:hyper}.

\subsection{Case Study (RQ5)}

Next, we visualize the learned ST embeddings and predicted ST data evolution as case studies. The showcases highlight two advantages of our proposed \model framework: $i)$ it can decompose the embedding space for latent factors that have an impact on ST data, and $ii)$ the predicted partial results corresponding to each factor can well explain the final overall predictions. Note that we use STGCN as the default base model in our case study for the balance of performance and efficiency.

\subsubsection{Decomposed Embedding Space for Latent Factors}

\begin{figure}[t]
    \centering
    % \begin{subfigure}[b]{0.45\columnwidth}
    %      \centering
    %      \includegraphics{figures/tsne-a.png}
    %      \caption{Embeddings learned by base model.}
    %      \label{fig:tsne-a}
    %  \end{subfigure}
    %  \hfill
    %  \begin{subfigure}[b]{0.45\columnwidth}
    %      \centering
    %      \includegraphics{figures/tsne-b.png}
    %      \caption{Embeddings learned by base model.}
    %      \label{fig:tsne-b}
    %  \end{subfigure}
    \subfigure[Embeddings learned by base model.]{\includegraphics[width=0.45\columnwidth]{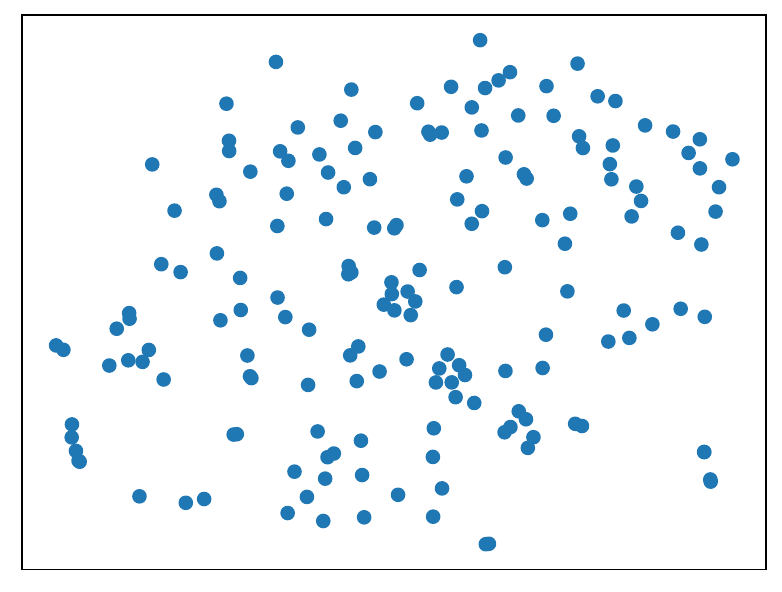}}
    \quad
    \subfigure[Embeddings learned by \model-enhanced model.]{\includegraphics[width=0.45\columnwidth]{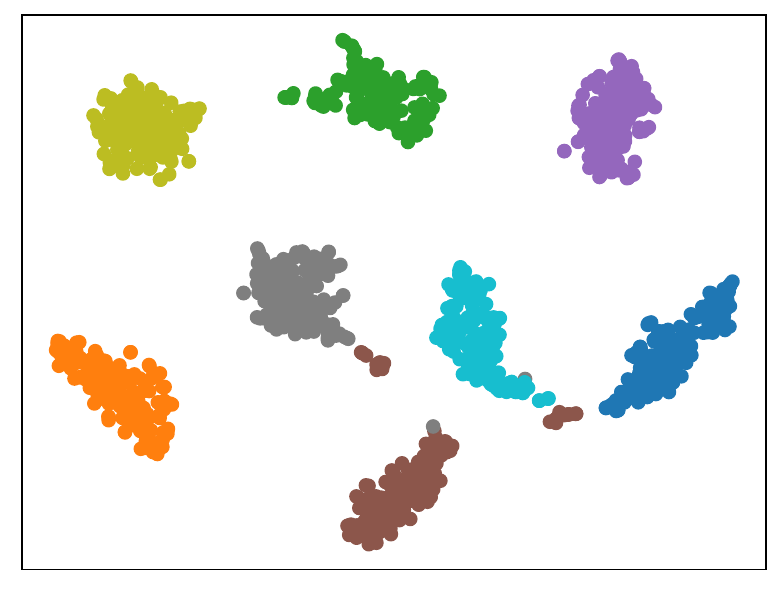}}
    % \subfigure[t-SNE visualization of the embeddings generated by enhanced STGCN on one of the subgraphs.]{\includegraphics[width=0.45\columnwidth]{figures/tsne-c.pdf}}
    % \caption{t-SNE visualizations on PEMSD8 dataset, including embeddings generated by both the basic STGCN and the enhanced STGCN, as well as embeddings generated by the enhanced STGCN on one of its subgraphs.}
    \vspace{-.2cm}
    \caption{Visualization of the learned ST embeddings.}
    \label{fig:tsne}\vspace{-.4cm}
\end{figure}

In the decomposed learning network (Sec.~\ref{ssec:dln}), there are several ST blocks that aim to learn partial ST data relevant to different latent factors. Each ST block first generates ST embeddings of the input data by an ST encoder, and then the embeddings are used for backcast and forecast. Only if the ST embeddings capture meaningful features for each latent factor, we can make accurate ST prediction. To explore whether our \model could learn quality embeddings, we visualize the learned embeddings of PEMDS8 by t-SNE. We randomly select a traffic ST data sample for visualization. The learned embeddings of the base model and the \model-enhanced version are shown in \fig~\ref{fig:tsne}(a) and (b), respectively. In these figures, each scatter denotes a node embedding of the traffic ST graph.

We can observe that the ST embeddings generated by the base model are relatively dispersed in the feature space, with no obvious patterns found. In contrast, the ST embeddings learned by our \model-enhanced model are significantly better categorized into 8 clusters. For better understanding, we use different colors to indicate the embedding of different decomposed subgraphs. We can find that each cluster corresponds to a subgraph that represents a certain latent factor. This is because the \model is designed for modeling the multi-factor ST prediction task. That is, each set of embeddings serves a downstream task that corresponds to a specific factor and therefore exhibits a degree of intrinsic similarity. 

Furthermore, we delve into the embeddings of each cluster. These enhanced embeddings, in contrast to those generated by the base model, tend to exhibit a stronger alignment along a \textit{one-dimensional axis} in a small area of the feature space (taking the orange cluster as an example). This alignment indicates a greater focus on modeling a singular factor within the input ST data, thereby demonstrating the necessity of our framework in decoupling the original ST prediction problem into multiple problems relevant to different latent factors.

\begin{figure}[t]
    \centering
    \subfigure[Prediction visualization. The solid blue line represents the ground truth (GT). The solid green line and solid red line represent the predictions of the base model and the \model-enhanced model, respectively. The dashed lines represent the predictions on 6 subgraphs within the enhanced model.]{
    \includegraphics[width=\columnwidth]{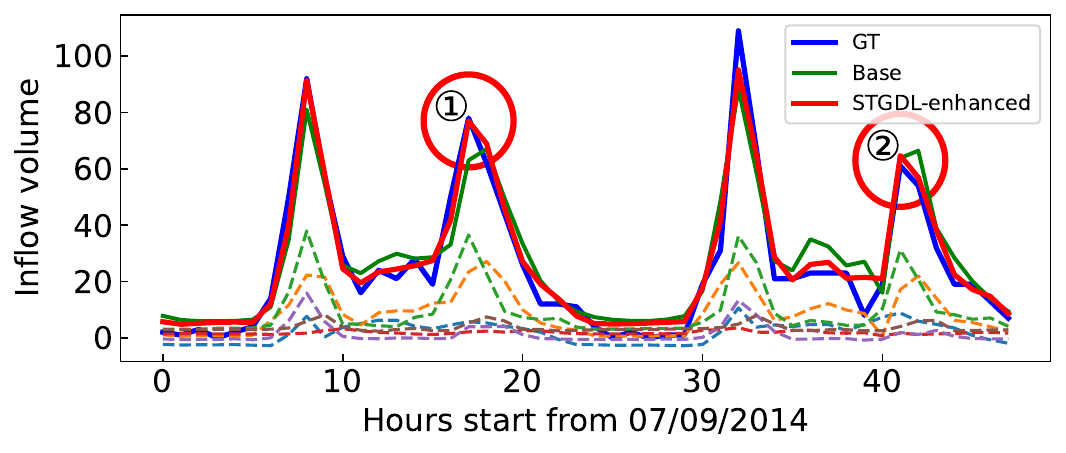}
    }
    \subfigure[Data distribution of predictions on 6 subgraphs.]{
    \includegraphics[width=\columnwidth]{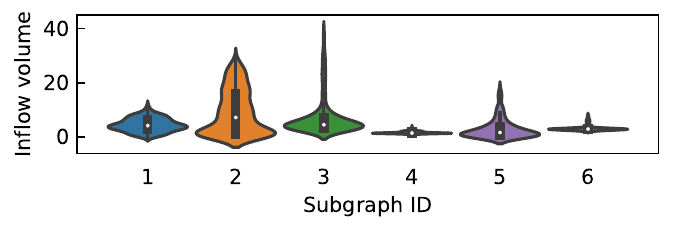}
    }%\vspace{-.2cm}
    \caption{Interpretability analysis of the prediction results.}
    \label{fig:pred}\vspace{-.3cm}
\end{figure}

\subsubsection{Interpretation for Traffic Prediction Results}

Here, we show the interpretability potential of our \model's predictions and its internal mechanism by visualizing the prediction results. 

From the model aspects, \fig~\ref{fig:pred}(a) illustrates a typical case on the NYCBike dataset, focusing on node 80, during a 48-hour period starting from 00:00 on 07/09/2014. We visualize the prediction results of \model-enhanced model and the base model. It can be observed that the enhanced model produces more accurate predictions, especially in peak hours, compared to the base model. 
To explain this, we plot the predicted data evolution on 6 subgraphs. It is these partial data components that contribute to the final ST prediction of the \model-enhanced model, so they can be used as an interpretation for model predictions.
Specifically, at point \textcircled{1}, the prediction of the base model and that of the subgraph (dashed orange line) exhibit similar \textit{bimodal} trends. However, due to the influence of other factors (such as factors corresponding to the dashed light green line), the actual data distribution should be \textit{unimodal}. Consequently, the predictions of the \model-enhanced model are more accurate than the basic counterpart. A similar interpretation example can also be found at point \textcircled{2}.

As for the internal mechanism, our \model aims to make multi-factor ST prediction by forecasting partial ST data under different factors separately. This means that the prediction results of different ST blocks are supposed to be distinct.
As shown in \fig~\ref{fig:pred}(b), the data distributions of ST predictions under different factors vary greatly. This indicates that our \model successfully captures the patterns of different latent factors.
We believe that guided by some expert knowledge or data, our \model can learn ST patterns that have real-world physical meaning.

\section{Related Work}\label{sec:related_work}

\paratitle{Spatio-Temporal Prediction} has drawn great attention in recent years due to its integral role in intelligent decision-making, scheduling, and management in smart cities~\cite{wang2023high, zheng2019urban, wang2017no}. To simultaneously model the temporal and spatial dependencies, most studies adopt deep learning techniques. For temporal dependency modeling, they employ the recurrent neural networks~\cite{liu2022msdr, bai2020adaptive, li2018dcrnn_traffic}, temporal convolutional networks~\cite{wu2019graph, wang2022traffic, ji2020interpretable}, or attention mechanism~\cite{yao2019revisiting}. 
For spatial dependency modeling, they utilize graph neural networks~\cite{yu2018spatio, guo2021learning, ji2023spatial, wu2020connecting} or attention mechanism over graphs~\cite{pan2022spatio, zhang2021traffic, wang2022traffic, zheng2020gman}. 
These models focus on designing a holistic model to capture all spatio-temporal patterns affected by multiple latent factors, making it challenging to accurately predict the evolution of spatio-temporal data. To overcome this limitation, we re-formulate the learning task as a multi-factor spatio-temporal prediction problem. We then incorporate graph decomposition to model multiple subgraphs relevant to different latent factors, effectively capturing the influence of multiple latent factors.

\paratitle{Graph Decomposition}, or graph partitioning, is a pivotal task in graph theory that involves decomposing complex input graphs into simpler subgraphs~\cite{bulucc2016recent}. The primary goal is to delve into the graph's organization and features more effectively, and the optimization objective is to divide the graph into balanced partitions while minimizing the cut across the partitions~\cite{feldmann2015balanced}. Due to its combinatorial nature, many approximate algorithms have been developed, including multilevel partitioning\cite{karypis1999multilevel}, spectral clustering\cite{ng2001spectral}, pagerank-based local partitioning\cite{andersen2006local},  simulated annealing \cite{kawamoto2018mean} and more. Recent advances in graph decomposition have seen the integration of deep learning techniques, including GAP\cite{nazi2019gap} which learns node embeddings and designs specific loss functions for the optimizing objective.
Graph partitioning is primarily applied in high-performance parallel computing. 
In the domain of spatio-temporal traffic forecasting, Ref.~\cite{mallick2020graph} partitions the graph into multiple parts to process DCRNN\cite{li2018dcrnn_traffic} on a large graph. However, in the aspect of decoupling spatio-temporal data, there is currently a lack of research and application of graph decomposition. To leverage the advantage of graph decomposition, we attempt to apply it to traffic prediction for modeling the impact of multiple factors. Unlike classical methods, the graph decomposition strategy we employ is not aimed solely at minimizing the cuts between subgraphs. Instead, it is integrated as a learnable module within our \model framework and constrained by regularization terms. 
% to constrain the distinctiveness among subgraphs and maintain the integrity of the original graph.

% To first to propose the application of learnable graph decomposition in the field of traffic forecasting. We capture different traffic patterns through graph decomposition and leverage the advantage of deep learning to automatically learn the decomposed subgraphs.

% \section{Experiments}

% 4 pages.

\section{Conclusion and Future Work}\label{sec:con}

This work investigated the spatio-temporal (ST) prediction problem from a multi-factor lens. We proposed a decomposed prediction strategy to address this problem theoretically. On top of that, we instantiated a novel ST graph decomposition learning framework called \model. It consists of two components: an automatic graph decomposition module and a decomposed learning network. The former decomposes the original graph structure into several subgraphs corresponding to different latent factors, while the latter learns the partial ST data on each subgraph separately and integrates them for the final prediction. Extensive experiments on four datasets showed that deep ST models integrating with our framework can achieve significantly better performance. Empirical studies confirmed the advantages of \model on the effectiveness and interpretability. In the future, we plan to allow STGDL to learn human-understandable ST data patterns by bringing in expert knowledge.

\section*{Acknowledgments}
\noindent Jingyuan Wang acknowledges support from the National Natural Science Foundation of China (No. 72222022, 72171013, 72242101).

\bibliographystyle{IEEEtran}
\bibliography{IEEEabrv,sample-base-v2}

% \begin{IEEEbiography}[{\includegraphics[width=1in,height=1.25in,clip,keepaspectratio]{figures/jiahao.jpg}}]{Jiahao Ji}
\vspace{-1cm}
\begin{IEEEbiography}[{\includegraphics[width=1in,height=1.25in,clip,keepaspectratio]{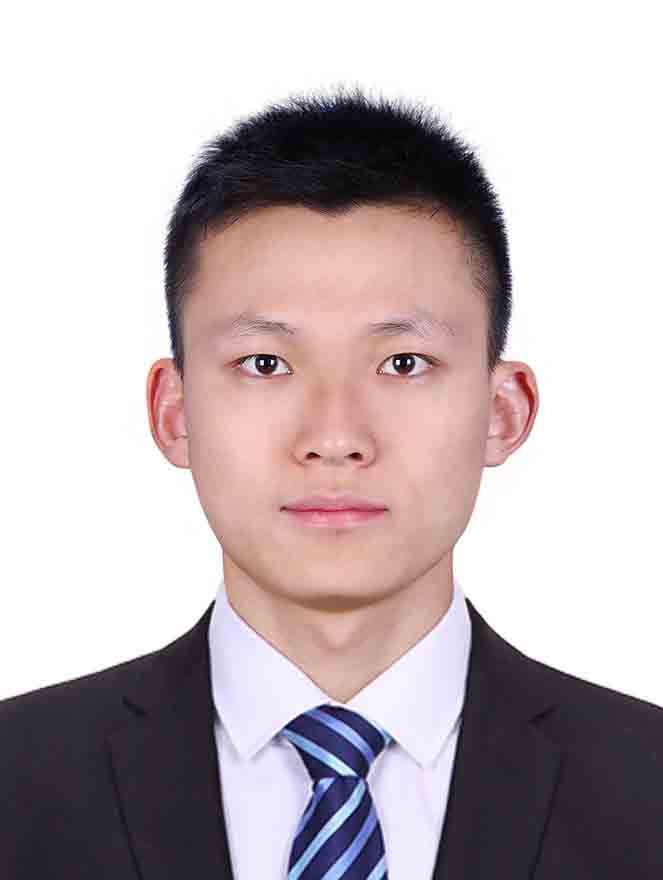}}]{Jiahao Ji}
is a Ph.D. candidate at the School of Computer Science and Engineering, Beihang University. He received his B.S from Beihang University in 2019. 
His research interests include spatio-temporal data mining, interpretable machine learning, and urban computing. 
\end{IEEEbiography}
\vspace{-1.1cm}

%%% biography section
\begin{IEEEbiography}[{\includegraphics[width=1in,height=1.25in,clip,keepaspectratio]{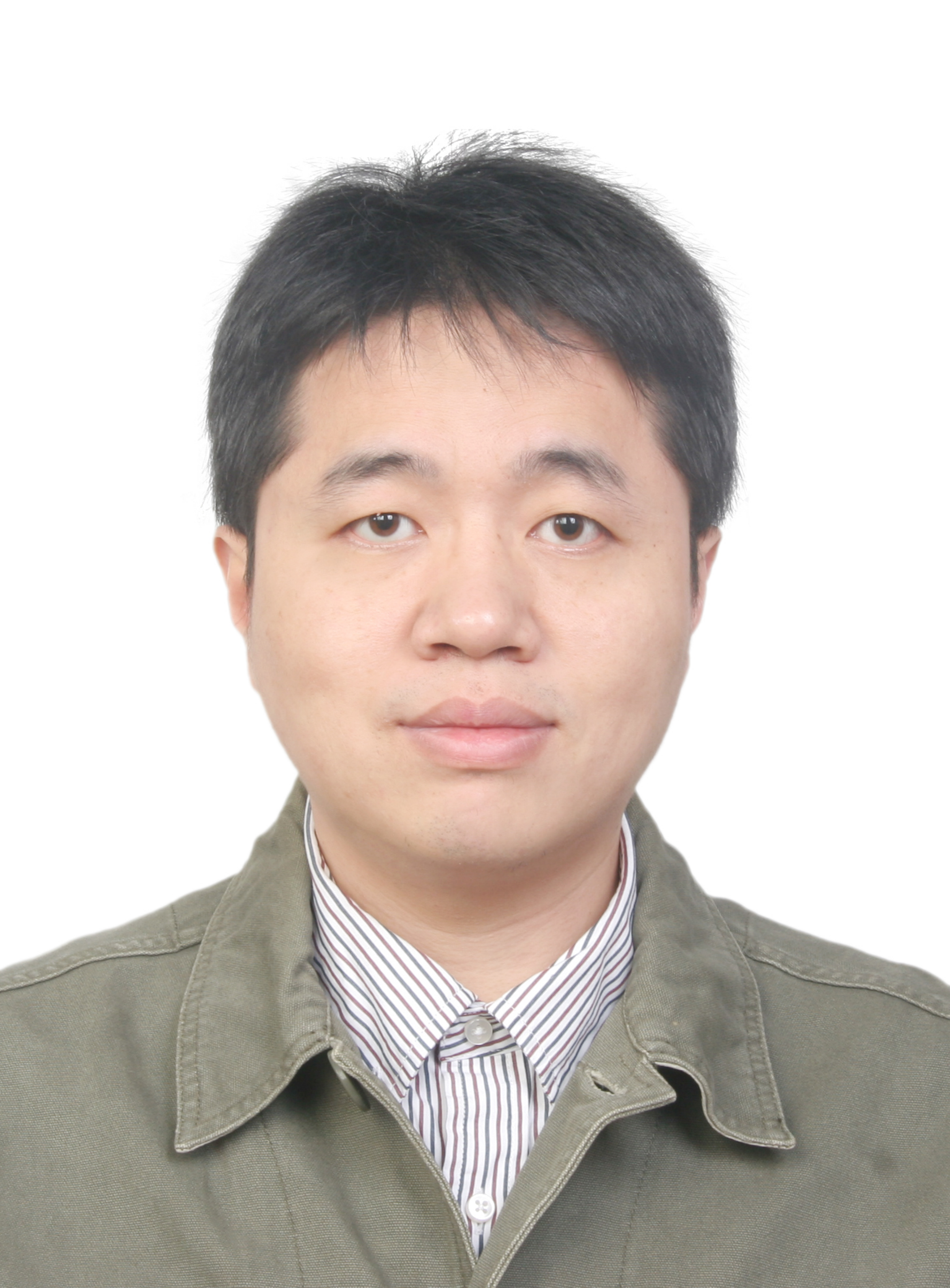}}]{Jingyuan Wang}
  received the Ph.D. degree from
  the Department of Computer Science and Technology,
  Tsinghua University. He is currently a Professor at School of Computer Science and Engineering, Beihang University. He
  is also the head of the Beihang Interest Group on
  SmartCity (BIGSCity), and Vice Director of the
  Beijing City Lab (BCL). His general area of research
  is data mining and machine learning,
  with special interests in smart cities and spatiotemporal
  data analytics.
\end{IEEEbiography}
\vspace{-1.1cm}

\begin{IEEEbiography}[{\includegraphics[width=1in,height=1.25in,clip,keepaspectratio]{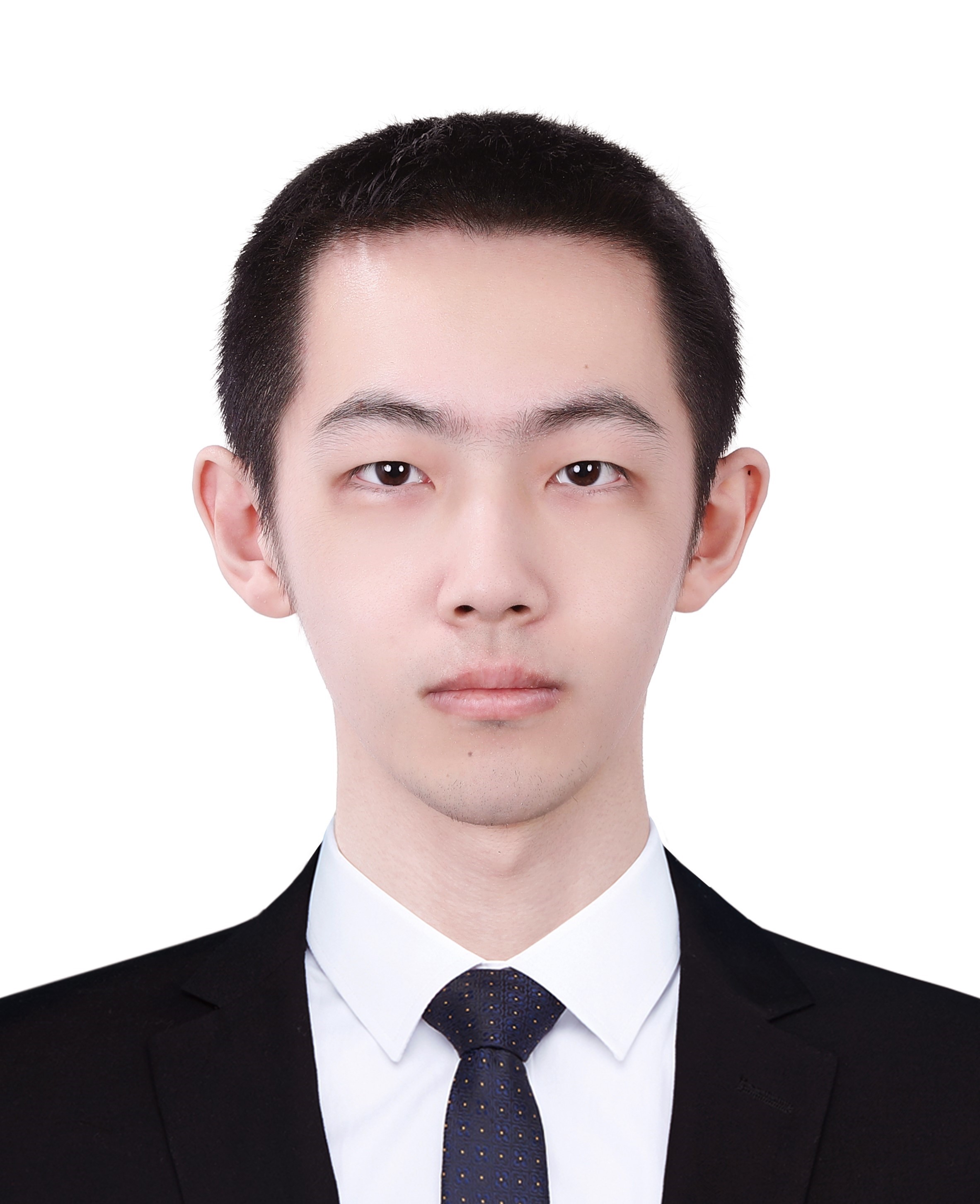}}]{Yu Mou}
is a graduate student at the School of Computer Science and Engineering, Beihang University. He received his B.S. from Beihang University in 2022. 
His research interests include road representation learning and spatio-temporal data mining. 
\end{IEEEbiography}
\vspace{-1.1cm}

% (S'11-M'15-SM'22)
\begin{IEEEbiography}[{\includegraphics[width=1in,height=1.25in,clip,keepaspectratio]{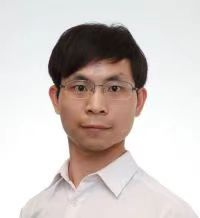}}]{Cheng Long}
is currently an Assistant Professor at the School of Computer Science and Engineering, Nanyang Technological University. He received his Ph.D. degree from the Hong Kong University of Science and Technology, Hong Kong, in 2015, and his BEng degree from South China University of Technology, China, in 2010. His research interests are broadly in data management and data mining. 
\end{IEEEbiography}

\appendices

\section{}\label{appx:lemmas}

\subsection{Proof of Lemma 1}
\begin{IEEEproof}
According to the graph decomposition step, it decomposes the original traffic graph $G$ into $K$ subgraphs, namely $\{G_k\}_{k=1}^{K}$. Consequently, the original ST prediction problem $P$ defined on $G$ is divided into $K$ subproblems, namely $\{P_k\}_{k=1}^{K}$. Each subproblem aims to predict the future ST data $\bm{X}^{(t+1)}_k$ corresponding to $G_k$.
Before delving into further proof, we give two necessary definitions.
\begin{mydef}[Problem Scale]\label{def:scale}
    Given a prediction problem $P$ on ST graph data $\mathcal{G}^{(t-T:t)} = (G, \bm{X}^{(t-T:t)})$ with graph structure $G = (\mathcal{V}, \mathcal{E}, \bm{A})$, we define the problem scale $B_P$ as the number of graph nodes, \ie $B_P = |\mathcal{V}|$.
\end{mydef}
\begin{mydef}[Problem Independence]
    Given two ST prediction problems $P_1$ and $P_2$, they are independent ($P_1 \perp P_2$) if and only if graph edge sets $\mathcal{E}_1$ and $\mathcal{E}_2$ of both problems are orthogonal, \ie $\mathcal{E}_1 \cap \mathcal{E}_2 = \emptyset$. 
\end{mydef}

For any subproblem $P_k, k \in [1, K]$, we know $G_k \subset G$, that is $|\mathcal{V}_k| < |\mathcal{V}|$. Thus, we can have $B_{P_k} < B_{P}$, meaning the scale of every subproblem is \textit{smaller} than that of the original problem. 
On the other hand, the subgraphs satisfy $\bigcap \mathcal{E}_k = \emptyset$, so each pair of graph edge sets satisfy
\begin{equation}\small
    \forall j, k \in [1, K], \mathcal{E}_j \cap \mathcal{E}_k = \emptyset \Rightarrow P_j \perp P_k.
\end{equation}
This means that any two subproblems are \textit{independent}.
\end{IEEEproof}

\subsection{Proof of Lemma 2}
\begin{IEEEproof}
According to \thm~\ref{thm:pred}, we know that $\forall k \in [1, K], e_k < e$. Here, $e_k$ is the error rate lower bound of solving the $k$-th subproblem $P_k$, while $e$ is that of directly solving the original problem $P$.
% According to \thm~\ref{thm:pred}, we know that the traffic flow predictability upper bound of the $k$-th subproblem ($\pi^{max}_k$) is greater than that of the original problem ($\pi^{max}$), \ie $\forall k \in[1, K], \pi^{max}_k > \pi^{max}$. 
% Because predictability tells prediction accuracy~\cite{song2010limits}, the error rate lower bound of a traffic prediction problem can be denoted by $e = 1 - \pi^{max}$. For the $k$-th traffic prediction subproblem $P_k$, we have $e_k = 1 - \pi^{max}_k$ and it satisfies
% \begin{equation}\label{eq:error_rate}
%     \forall k \in [1, K], e_k < e,
% \end{equation}
% where $e$ is the error rate lower bound of directly solving the original traffic prediction problem $P$. 
From the data generation perspective, the partial ST data relevant to the $k$-th subproblem can be considered as samples of a random variable $X_k$. We use $\sigma^2_k$ to denote the variance of $X_k$. Then, we analyze the prediction error lower bound of the $k$-th subproblem, namely $ELBO_k$. The error comes from two cases: $i)$ predictions are correct and $ii)$ predictions are incorrect. Thus, the $ELBO_k$ can be computed by the expectation of errors in both cases:
\begin{equation}\label{eq:elbo_k}\small
    ELBO_k = e_k \cdot r_k^{0} + (1-e_k) \cdot r_k^{1}. 
\end{equation}
where $e_k$ is the error rate lower bound indicating the minimal probability of making incorrect predictions, and $r_k^{0}$ is the error expectation when the predictions are incorrect. $r_k^{1}$ is that when the predictions are correct.

We know $r_k^{1} = 0$. As for $r_k^{0}$, it can be calculated by the error between every possible prediction and ground truth:
\begin{equation}\label{eq:err_exp}\small
    r_k^{0} = \sum_i p_i^{(k)} \sum_j p_j^{(k)} \left(x_j^{(k)} - x_i^{(k)}\right)^2,
\end{equation}
where $x_i^{(k)}$ is a ground truth with probability $p_i^{(k)}$, and $x_j^{(k)}$ is a prediction with probability $p_j^{(k)}$. Solving \equ~\eqref{eq:err_exp} is equivalent to finding the solution of $\mathbb{E}[(X-Y)^2]$ with constraint $i)$ $X \perp Y$ and $i)$ variables $X, Y$ are the instance of the same random variable. Since
\begin{equation}\small
\begin{aligned}
    \mathbb{E}[(X-Y)^2] =& \mathbb{E}[X^2-2XY+Y^2]\\
    =& \mathbb{E}[X^2] - 2\mathbb{E}[XY] + \mathbb{E}[Y^2]\\
    =& \mathbb{E}[X^2] - 2\mathbb{E}[X]\mathbb{E}[Y] +\mathbb{E}[Y^2],\\
\end{aligned}
\end{equation}
we substitute $X, Y$ with a unified variable $Z$ and have
\begin{equation}\small
\begin{aligned}
    \mathbb{E}[(X-Y)^2] =& \mathbb{E}[Z^2] - 2\mathbb{E}[Z]\mathbb{E}[Z] +\mathbb{E}[Z^2]\\
    =& 2(\mathbb{E}[Z^2] - \mathbb{E}^2[Z]) = 2D(Z),\\
\end{aligned}
\end{equation}
where $D(Z)$ is the variance of random variable $Z$.
Through the above derivations, we know that $r_k^{0} = 2\sigma_k^2$. Taking $r_k^{1}$ and $r_k^{0}$ into \equ~\eqref{eq:elbo_k}, we can have
\begin{equation}\label{eq:elbo_k_ana}\small
    ELBO_k = e_k \cdot 2\sigma_k^2 + (1-e_k) \cdot 0 =  2 e_k \sigma_k^2.
\end{equation}
That is, the error lower
bound of the $k$-th subproblem is $2 e_k \sigma_k^2$.
\end{IEEEproof}

\subsection{Proof of Lemma 3}
\begin{IEEEproof}
By combining the solutions to all subproblems, we can have an integrated solution for the original problem. From Lemma~\ref{lem:divide}, the subproblems are independent of each other. This means that the covariance of any two subproblems is zero, thereby making the covariance of the relevant random variables zero. Therefore, the error lower bound of the combined solution ($E_d$) can be computed by the summation of lower bounds of all subproblems, \ie $E_d = \sum_{k=1}^{K} ELBO_k = \sum_{k=1}^{K} 2 e_k \sigma_k^2$.
% \begin{equation}\label{eq:elbo_dac}\small
%     E_d = \sum_{k=1}^{K} ELBO_k = \sum_{k=1}^{K} 2 e_k \sigma_k^2.
% \end{equation}
% The proof finishes here. 
\end{IEEEproof}

% \section{Analysis of Traffic Flow Predictability}\label{appx:predictability}
\section{}\label{appx:predictability}

\subsection{Proof of Error Rate Lower Bound}
% 定理2：可预测性的关系
\begin{theorem}\label{thm:pred}
    In the decomposed prediction strategy, the error rate lower bound of any subproblem ($e_k$) is smaller than that of the original problem ($e$), \ie $\forall k \in [1, K], e_k < e$. 
\end{theorem}

Here, the error rate lower bound ($e$) describes the minimal probability of making incorrect predictions. It can be computed by the predictability upper bound ($\pi^{max}$)~\cite{song2010limits} via 
\begin{equation}\label{eq:err_rate_pred}\small
    e = 1 - \pi^{max}.
\end{equation}
Predictability refers to the extent to which future data can be accurately predicted with only past observations. The data to be predicted in this paper are ST data. They are naturally time series, allowing us to use tools from information theory, including entropy~\cite{brabazon2008natural} and Fano's inequality~\cite{cover2006elements}, to measure the predictability.
% It can be measured by the information entropy~\cite{Brabazon2008}.
Next, we prove \thm~\ref{thm:pred} through predictability and information entropy.
% In order to prove \thm~\ref{thm:pred}, we introduce two lemmas.

% \begin{lemma}\label{lem:ent}
%     The entropy of ST data distribution under any single factor ($S_k$) is smaller than that under multiple factors ($S$), \ie $\forall k\in[1, K], S_k < S$. 
% \end{lemma}

% \begin{lemma}\label{lem:pred}
%     The predictability upper bound of ST data under any single factor ($\pi_k^{max}$) is greater than that under multiple factors ($\pi^{max}$), \ie $\forall k\in[1, K], \pi_k^{max} > \pi^{max}$. 
% \end{lemma}

% Then, we give the proof of \thm~\ref{thm:pred}.
\begin{IEEEproof}
According to Fano’s inequality~\cite{cover2006elements}, we can compute the predictability upper bound ($\pi^{max}$) by solving
\begin{equation}\label{eq:pi_max}
\begin{aligned}
    S =& H(\pi^{max}) + (1-\pi^{max}) \log(M-1)\\
    =& -\pi^{max}\log(\pi^{max})-(1-\pi^{max})\log(1-\pi^{max})\\
    &+ (1-\pi^{max}) \log(M-1).\\
\end{aligned}
\end{equation}
Here, $S$ is the entropy of data distribution. $H(\pi^{max})$ is the entropy of $\pi^{max}$. $M$ is the data state number, and it will be fixed once the data is given. 

We treat $S$ as a function $\pi^{max}$. Taking the first-order derivation of $S$ over $\pi^{max}$ yields
\begin{equation}\label{eq:first_order}
     S' = -\log(\pi^{max}) + \log(1-\pi^{max}) - \log(M-1).
\end{equation}
Then, taking the second-order derivation of $S$ yields
\begin{equation}\small
    S'' = \frac{1}{\pi^{max}(1-\pi^{max})}.
\end{equation}
Since $\pi^{max} \in [0, 1]$, the inequality $S'' < 0$ always holds. This indicates that $S$ is a concave function on $\pi^{max}$ and $S$ has a maximum value. 

Let $S' = 0$ to obtain the maximum value point as $\pi^{max} = 1/M$. That is, when $\pi^{max} > 1/M$, $S$ is a monotonically decreasing function of $\pi^{max}$. Oppositely, when $\pi^{max} < 1/M$, $S$ is a monotonically increasing function of $\pi^{max}$. 
Since it was found that $\pi^{max} > 1/M$ usually holds for ST data such as human mobility data~\cite{song2010limits, wang2015predictability}, $S$ can generally be considered as a monotonically decreasing function of $\pi^{max}$. This means that the larger $S$ is, the smaller the corresponding $\pi^{max}$ is. 

Next, it is sufficient to compare entropy($S$) corresponding to the subproblems and the original problem to obtain a relation for predictability upper bound ($\pi^{max}$), and thus for the error rate lower bound ($e$). To this end, we introduce the following lemma.

\begin{lemma}\label{lem:ent}
    The entropy of ST data distribution under any single factor ($S_k$) is smaller than that under multiple factors ($S$), \ie $\forall k\in[1, K], S_k < S$. 
\end{lemma}

According to Lemma~\ref{lem:ent}, $S_k < S$ holds for any $k$. Since $S$ monotonically decrease with $\pi^{max}$, we have $\pi_k^{max} > \pi^{max}$ for any $k$. Bringing the inequality into \equ~\eqref{eq:err_rate_pred}, we have $\forall k \in [1, K], e_k < e$.
\end{IEEEproof}

\subsection{Proof of Lemma 4}

\begin{IEEEproof}
    Before delving into the proof, we clarify some terms. We use $X_k$ ($k \in [1, K]$) to denote the random variable relevant to ST data under the $k$-th factor. $S_k$ is the corresponding entropy. Since the data under multiple factors is a mixture of data under all single factors, $S$ is the joint entropy of all random variables and can be defined as
    \begin{equation}\label{eq:mf_ent}\small
        S = H(X_1, X_2, \dots, X_K) = \sum_{k=1}^{K} H(X_k|X_{k-1}, \dots, X_1),
    \end{equation}
    where $H(X_k|X_{k-1}, \dots, X_1)$ is the conditional entropy. 

    There are two steps to prove this lemma: $i)$ proving $S_1 < S$. $ii)$ extending the conclusion of $S_1$ to $S_k$. 
    For step $i)$, we expand the summation term in \equ~\eqref{eq:mf_ent}:
    \begin{equation}\label{eq:ent_ss1}\small
    \begin{split}
        S &=H(X_1, X_2, \dots, X_K)\\
        &= H(X_1) + H(X_2| X_1) + \dots + H(X_K|X_{K-1},\dots,X_1)\\
        &\ge H(X_1) = S_1.\\
    \end{split}
    \end{equation}
    The last step is derived from the non-negativity of conditional entropy. It means that in step $ii)$, each term after $H(X_1)$, such as $H(X_2|X_1)$, is non-negative. 

    For step $ii)$, according to the symmetry of entropy, entropy should be unchanged if the outcomes $X_k$ are re-ordered. Therefore, we can rewrite the expression of $S$ as
    \begin{equation}\small
       S = (X_k, X_1, \dots, X_{k-1}, X_{k+1}, \dots, X_K).
    \end{equation}
    Here, we adjust $X_k$ from the $k$-th position to the first position. This operation does not change the value of entropy. By using the rules of \equ~\eqref{eq:ent_ss1}, we can obtain
    \begin{equation}
        S_k \le S, k \in [1, K].
    \end{equation}
    The equal sign is obtained when ST data of all factors are the same. This is nearly impossible in real-world scenarios, so we remove the equal sign and reach the final conclusion that $\forall k \in [1, K], S_k < S$.
\end{IEEEproof}

\end{document}